%% file: latex/main.tex
\documentclass[11pt]{article}

\usepackage[final]{acl}

\usepackage{times}
\usepackage{latexsym}
\usepackage{algorithm}
\usepackage{algpseudocode}
\usepackage[most]{tcolorbox}
\usepackage{xcolor}
\usepackage{subcaption}
\usepackage{booktabs}
\usepackage{array}
\usepackage{arydshln}
\usepackage{fontawesome5}

\definecolor{codegreen}{rgb}{0.0, 0.3, 0.0} 
\definecolor{codepurple}{rgb}{0.65, 0.15, 0.32} 

\definecolor{methodblue}{RGB}{35,85,160}
\definecolor{methodbluelight}{RGB}{246,249,255}
\definecolor{methodgray}{RGB}{245,245,247}

\tcbset{
  methodbox/.style={
    enhanced,
    colback=methodbluelight,
    colframe=methodblue,
    coltitle=white,
    colbacktitle=methodblue,
    fonttitle=\bfseries\small,
    boxrule=0.7pt,
    arc=2mm,
    left=1.5mm,
    right=1.5mm,
    top=1mm,
    bottom=1mm,
    toptitle=0.7mm,
    bottomtitle=0.7mm,
    attach boxed title to top left={xshift=1.5mm,yshift=-1.5mm},
    boxed title style={
      colback=methodblue,
      colframe=methodblue,
      arc=1.5mm,
      boxrule=0pt,
      left=1mm,
      right=1mm,
      top=0.5mm,
      bottom=0.5mm
    }
  }
}
\newtcolorbox{methodalgobox}[2][]{
  methodbox,
  title={#2},
  label={#1}
}
\usepackage{caption}
\usepackage[T1]{fontenc}

\usepackage[utf8]{inputenc}

\usepackage{microtype}

\usepackage{inconsolata}
\usepackage{xcolor}
\usepackage{amsmath}
\usepackage{amssymb}
\usepackage{booktabs}

\usepackage{graphicx}
\newcommand{\mask}{\texttt{[MASK]}}
\newcommand{\sage}{\textsc{AXON}}
\newcommand{\method}{\text{AXON}}

\title{Supportive Token Revealing for Fast Diffusion Language Model Decoding }

\author{
  \textbf{Giries Abu Ayoub\textsuperscript{1}},
  \textbf{Mario Barbara\textsuperscript{1}},
  \textbf{Lluís Pastor-Pérez\textsuperscript{2}},
  \textbf{Tanja Bien\textsuperscript{2}},
  \\
  \textbf{Aneesh Barthakur\textsuperscript{2}},
  \textbf{Alaa Maalouf\textsuperscript{1}},
  \textbf{Loay Mualem\textsuperscript{2,3}}\thanks{Corresponding author}
  \\ \\
  \textsuperscript{1}Department of Computer Science, University of Haifa \\
  \textsuperscript{2}Institute for AI, University of Stuttgart
    \textsuperscript{3}IMPRS-IS
  \\
\small{
\textbf{Corresponding authors: }
\href{mailto:jerryabuayob@gmail.com}{jerryabuayob@gmail.com}
and
\href{mailto:loaymua@gmail.com}{loaymua@gmail.com}
}
}

\begin{document}
\maketitle
\begin{abstract}
Discrete diffusion language models can generate text efficiently by updating multiple masked positions in parallel, but this parallelism introduces a quality-latency trade-off. Aggressive decoding may commit mutually dependent tokens too early, while conservative decoding requires many denoising steps. Existing methods address this tension by deciding which tokens are safe to reveal using confidence or dependency criteria. However, avoiding unsafe commits does not necessarily make the remaining masked sequence easy to decode, since uncertain tokens may depend on masked tokens, creating a bottleneck for denoising steps. We propose \method{}, a training-free module that can be added on top of existing parallel decoding strategies for diffusion language models. Rather than replacing the base decoder, \method{} monitors the remaining uncertain masked tokens and intervenes only when their current state suggests that additional context is needed. It then shifts the criterion from which tokens are safest to reveal to which confident reveals would best support later denoising. \method{} selects anchors, confident masked tokens that uncertain positions attend to, using attention, uncertainty, and confidence signals. Experiments on reasoning and code-generation benchmarks across multiple diffusion language models show that \method{} improves the quality-latency trade-off of existing parallel decoders, often reducing the number of function evaluations while maintaining or improving accuracy.

\faGlobe \space \href{https://jerryaa98.github.io/AXON-website/}{\underline{https://jerryaa98.github.io/AXON-website/}}

\end{abstract}
\input{latex/Loay_sections/Intro}
\input{latex/shortened_sections/related-shortened}
\input{latex/Loay_sections/Background2}

\input{latex/Loay_sections/Method}
\input{latex/experiments}
\input{latex/ack}

\clearpage
\bibliography{latex/custom}
\clearpage
\input{latex/appendix}

\end{document}

%% file: latex/Loay_sections/Intro.tex
\section{Introduction}

\begin{figure*}[t!]

\centering
\includegraphics[width=0.9\linewidth]{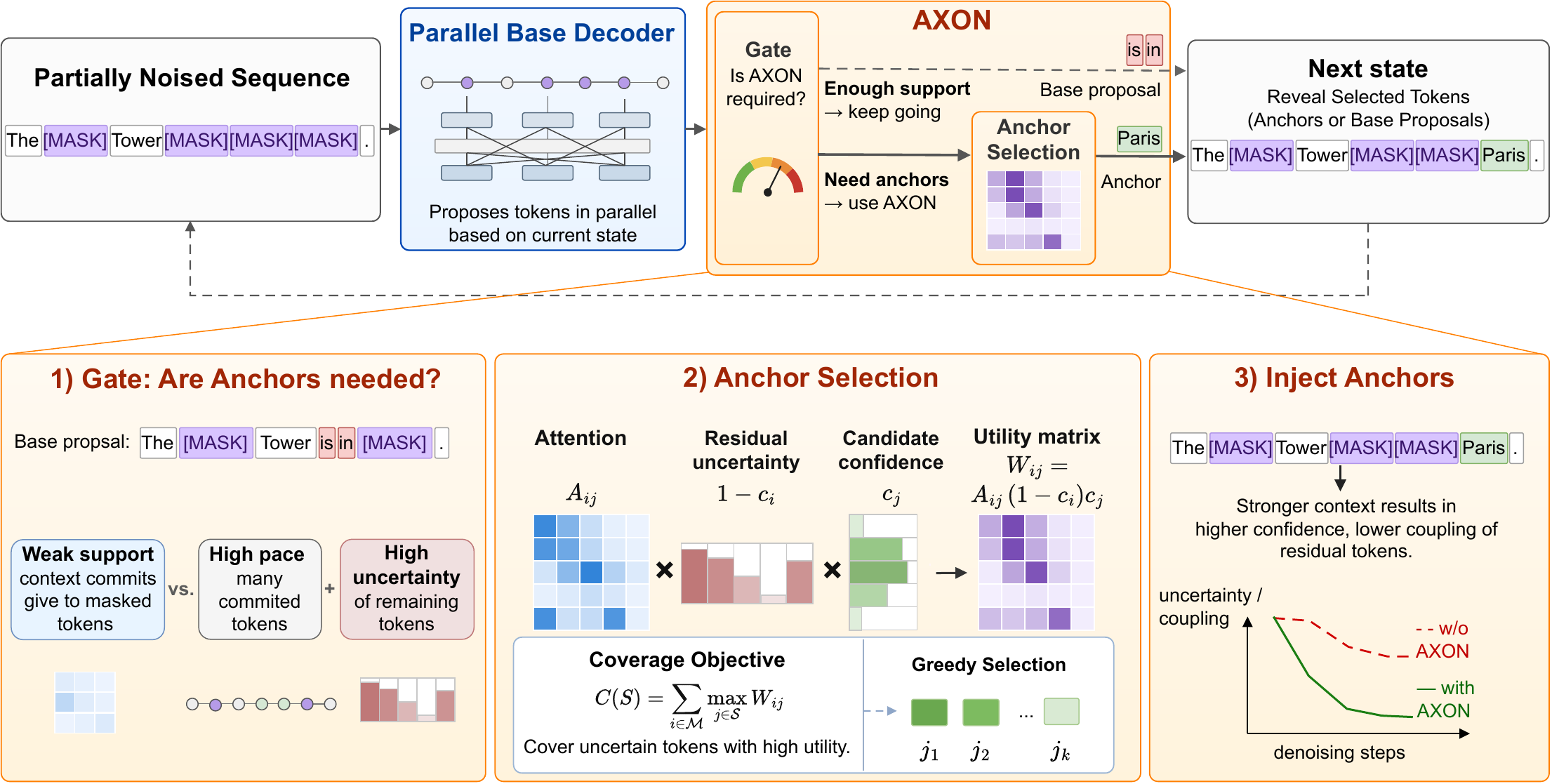}
\caption{
Overview of \method. A fast parallel decoder proposes tokens from the current masked state. AXON uses a lightweight gate to detect insufficient contextual support, selects influential anchors using attention, uncertainty, and confidence signals, and reveals them as context for the next denoising step.
\label{fig:overview_axon}
}
\end{figure*}
Autoregressive language models~\cite{radford2019ar1,brown2020ar2} generate text one token at a time, which makes their inference inherently sequential. Diffusion large language models (dLLMs)~\cite{shi2024simplified,sahoo2024dllm1,li2022dlmm2,gong2022dlmm3,louddiscretedlmm4} offer a different generation paradigm: they start from a masked sequence and iteratively denoise it, predicting all masked positions at each step. This bidirectional refinement process makes dLLMs attractive for low-latency text generation since multiple positions can be updated in parallel. Recent dLLMs such as LLaDA~\cite{nie2026llada} and Dream~\cite{ye2025dream} demonstrate that this paradigm can produce competitive text generations while exposing a natural source of parallelism that is absent in autoregressive left-to-right decoding.
However, unmasking arbitrary tokens simultaneously can produce inconsistent configurations as tokens may be coupled through syntax, semantics, arithmetic constraints, or multi-token entities. Existing training-free parallel-decoding methods mainly address this issue by deciding which tokens can be safely committed together. Confidence-based methods~\cite{wu2025fastdllm,ben2026ebsampler,kim2026klass,shu2026dcd} reveal tokens whose marginal probabilities are sufficiently high, implicitly assuming that high confidence tokens are both reliable and uncoupled/mutually independent. More recent dependency-aware methods~\cite{luo2026dawn,ringel2026dependency,zhou2026attention,kim2026dependency} use attention or distributional similarity to estimate interactions between masked positions and to avoid simultaneously committing strongly coupled tokens. However, even when unsafe commits are avoided, the remaining masked positions may still lack sufficient visible context for reliable parallel decoding. This creates a bottleneck where decoding becomes increasingly conservative or stalls altogether. 

In this work, we take a complementary view of fast dLLM decoding. Instead of only asking which tokens are safe to reveal immediately, we ask which masked tokens would provide the most useful context for the remaining uncertain positions if revealed early. Some tokens are already easy to commit because they are confident and weakly coupled to others. Other tokens are influential because many uncertain positions attend to them or depend on them. Revealing such tokens can improve contextual support for later denoising steps and reduce decoding bottlenecks.

Motivated by this view, we propose \method{}\footnote{We call our method \method{} by analogy to axons in neural systems: it reveals a few high-impact tokens that transmit useful context to uncertain positions during denoising.}, a training-free module that can be added on top of existing parallel decoding strategies. \method{} treats influential-token revealing as a small set-selection problem. It scores candidate tokens using attention, uncertainty, and confidence signals, and selects a small non-redundant subset through a coverage-based objective. The coverage view naturally leads to a submodular function objective, since once an uncertain position is already supported by one selected token (i.e. covered), another token supporting the same position should contribute less additional value. Since this selection is not needed at every step, \method{} uses a lightweight gate that activates it only when the current masked tokens appears under-supported. Figure~\ref{fig:overview_axon} provides an overview of this process.

We evaluate \method{} on reasoning and code-generation benchmarks across multiple diffusion language models and parallel decoders. The results show that \method{} improves the quality-latency trade-off over strong baselines. Explicitly, our contributions are as follows :
\begin{itemize}
    \item We introduce a complementary view of fast dLLM decoding: beyond selecting tokens that are safe to commit, a decoder can reveal informative masked tokens that provide useful context for the remaining positions.

    \item We propose \method{}, a training-free module for existing parallel dLLM decoders. \method{} intervenes only when the current masked sequence appears under-supported. It first applies a gate to detect such states, then identifies confident masked tokens that are strongly attended to by uncertain positions, and finally selects a non-redundant subset of these anchors through a coverage-based objective. Revealing the selected anchors provides targeted context to the remaining masked positions without modifying the underlying model.

    \item We evaluate \method{} across multiple backbones, reasoning and code-generation benchmarks, and several families of parallel decoders, including confidence-based, locality-aware, and dependency-aware strategies. Across these settings, \method{} improves the quality-latency trade-off while keeping the backbone frozen.
\end{itemize}

%% file: latex/shortened_sections/related-shortened.tex
\section{Related Work}
\label{sec:related}
Fast dLLM inference methods primarily differ in how masked positions are selected and revealed during denoising. In the sequel, we first review discrete dLLMs, followed by training-free decoding methods for frozen backbones and dependency-aware parallel decoding methods. 

\noindent
\textbf{Discrete dLLMs.} Extending denoising diffusion to discrete spaces, discrete dLLMs parameterize the reverse process through per-token marginal predictions~\citep{shi2024simplified,hoogeboom2021argmax,austin2021structured,sahoo2024dllm1,muller2026support}. Recent large-scale masked dLLMs like LLaDA~\citep{nie2026llada,bie2025llada2,bie2026llada2} and Dream~\citep{ye2025dream} offer a competitive alternative to autoregressive generation.

\noindent
\textbf{Fast decoding and commit expansion for frozen dLLMs.} A way to accelerate decoding is to predict multiple tokens per
model call. In autoregressive models, speculative decoding and related
verification-based methods propose multiple future tokens and validate
them in parallel with the target model~\citep{leviathan2023speculative}.
For dLLMs, the challenge is to decide which masked positions can be
reliably committed together. Early approaches
make this choice using local uncertainty measures such as entropy,
margin, or confidence~\citep{ghazvininejad2019mask}. Fast-dLLM~\citep{wu2025fastdllm}
combines block-wise Key-Value caching with a rule that commits positions whose marginal confidence exceeds a threshold, using high confidence as a proxy for reliable parallel commitment. EB-Sampler~\citep{ben2026ebsampler} uses an entropy bound to adaptively choose which and how many tokens to unmask at each step. KLASS~\citep{kim2026klass} requires the
token's distribution to be both confident and stable across successive denoising steps. LocalLeap~\citep{kong2025localleap} expands commits around confident anchors by relaxing nearby positions 
within a local window when local predictions appear reliable. These methods are lightweight and
training-free, and they provide different criteria for expanding or
accepting commits in a frozen dLLM. 

\noindent
\textbf{Dependency-aware parallel decoding.}
Recent work has shown that aggressive parallel decoding can suffer from joint inconsistency, where tokens that look plausible individually become incompatible when committed together \citep{wu2025fastdllm,ben2026ebsampler,zhang2026generation}. Dependency-aware methods address this by estimating interactions between masked positions during decoding. APD~\citep{israel2026adp} uses an auxiliary autoregressive model to estimate dependencies between tokens committed in parallel. Similarly, DEMASK~\citep{ringel2026dependency} uses a lightweight pairwise dependency predictor trained on hidden states.
Instead of training a model, other methods utilize the already available attention maps to estimate dependencies.
Attn-Sampler~\citep{zhou2026attention} uses attention column sums as a global importance score for ordering commits. DAPD~\citep{kim2026dependency} builds an attention-induced Markov random field and picks per-step commits as a greedy independent set. DAWN~\citep{luo2026dawn} constructs an attention dependency graph and uses committed anchors to support decoding of related positions, with conflict-aware scheduling that avoids jointly updating strongly coupled positions.

These methods highlight the role of dependency structure in safe
parallel commitment, but they primarily use it to decide which positions
should or should not be updated together. \method{} instead focuses on
the complementary problem of anchor selection: identifying confident
reveals that are not only reliable, but also informative for
the remaining uncertain positions. Additional related work can be found in Appendix~\ref{sec:app_related}  .



%% file: latex/Loay_sections/Background2.tex
\section{Background}
\label{sec:background}

Diffusion language models generate a sequence by gradually resolving masked
positions. 
Let $X$ denote the prompt and let
$y=(y_1,\ldots,y_L)$ denote the response to be generated, where each
$y_i$ belongs to a discrete vocabulary $\mathcal{V}$. Decoding starts
from a fully masked response,
$y^{(0)}=(\mask{},\ldots,\mask{}),$
and proceeds through a sequence of denoising steps. At step $t$, the
current response $y^{(t)}$ contains both revealed tokens and still-masked
positions. We denote the set of masked positions by
$M^{(t)}=\{i\in\{1,\ldots,L\}: y_i^{(t)}=\mask{}\}.$
For every $i\in M^{(t)}$, the model predicts a categorical distribution
over the vocabulary,
$p_\theta(y_i=v\mid X,y^{(t)}), v\in\mathcal{V},$
and we define its confidence as the probability assigned to the most likely non-mask token.
A decoding rule then chooses a subset of masked positions
$R^{(t)}\subseteq M^{(t)}$ to reveal. The selected positions are filled
with their top-1 predictions and remain visible in the following
denoising steps. Thus, the main inference-time decision is how to choose
$R^{(t)}$ at each step.

\noindent
\textbf{Parallel decoding.} The advantage of dLLMs comes from the ability to reveal more than one
position per forward pass. Revealing many positions reduces the number of
denoising steps, but it can also introduce errors because the model gives
per-position marginal predictions while the selected positions may be
interdependent.  A parallel decoder effectively relies on the approximation $\;p_\theta(y_R\mid X,y^{(t)})
\approx
\prod_{i\in R} p_\theta(y_i\mid X,y^{(t)}),$
for the selected set $R$. This approximation is reliable when the
positions in $R$ are close to conditionally independent under the current
context. In practice, masked tokens can depend on one another through
syntax, semantics, arithmetic constraints, formatting, or multi-token
entities. Thus, tokens that look plausible individually may form
an inconsistent joint assignment when revealed together.

%% file: latex/Loay_sections/Method.tex
\textbf{Notation.}
Let $X$ denotes the conditioning prompt. At a given step $t$, let $y^{(t)}$ the set of tokens already revealed, $B^{(t)}$ denote the set of all token positions in the current decoding block, $M^{(t)}$ be the set of positions that are still masked, and let $U^{(t)} \subseteq M^{(t)}$ be the set of (\emph{uncertain}) positions whose predicted confidence falls below a threshold. Let $S^{(t)}_{\text{base}}$ denote the set of tokens that is proposed by the base decoder, and $S^{\star(t)}$ be the set of (final) tokens returned by \method{}. We denote by $c_i^{(t)} \in [0,1]$ the predicted confidence at position $i$, by $A_{ij}^{(t)}$ the attention weight that position $i$ assigns to position $j$, aggregated across all attention layers. Finally $\alpha \in [0,1]$ denotes the gate hyperparameter.

\section{Methodology} 
\label{sec:method}

\begin{figure*}[t]
\centering
\includegraphics[width=0.85\linewidth]{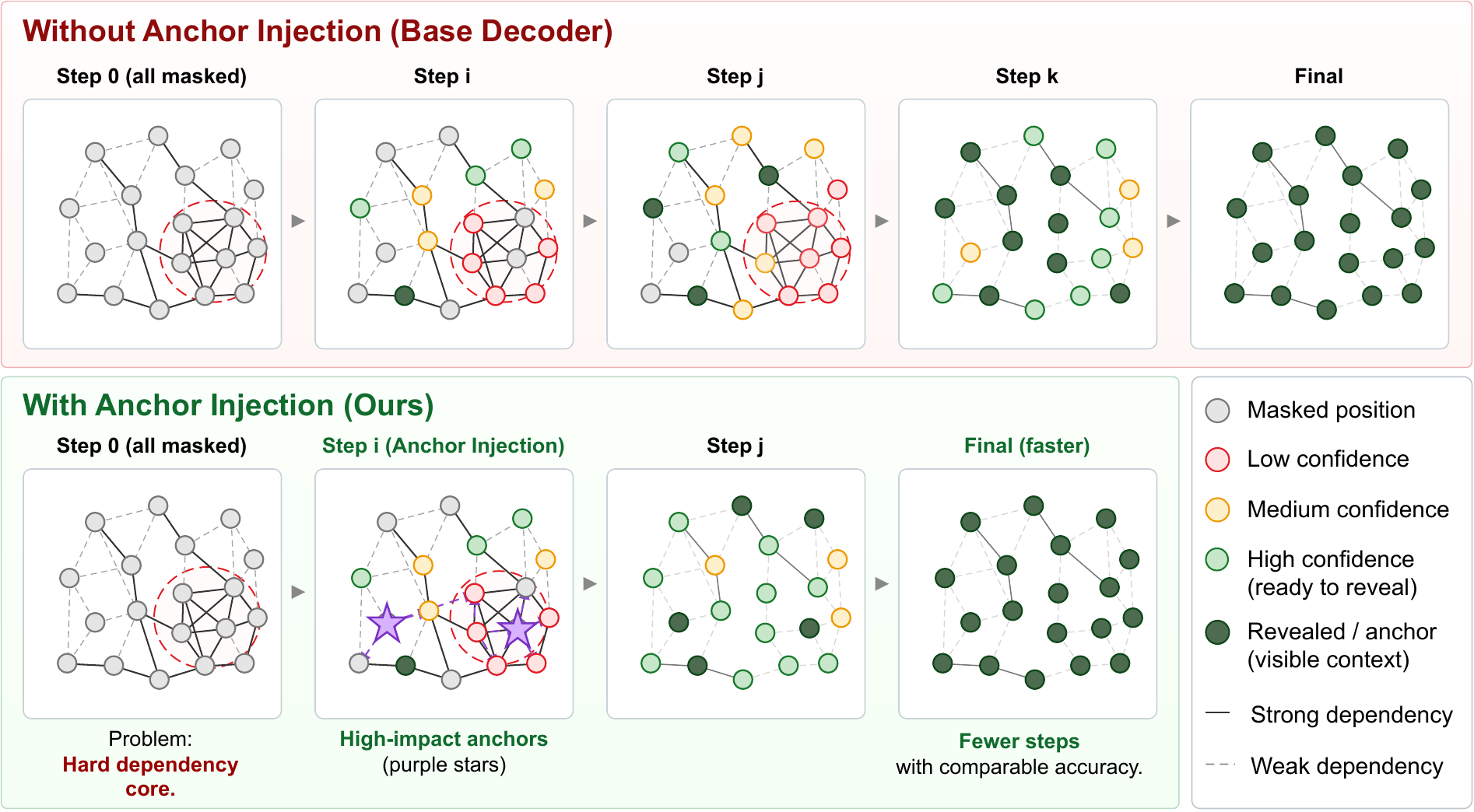}
\caption{\textbf{Effect of anchor injection in \method{}.} Without extra context, the base decoder struggles with a hard dependency core (red circle), requiring more denoising steps. When an under-supported state is detected, \method{} injects informative anchors (purple stars) as visible context. This breaks down unresolved dependencies, simplifying subsequent predictions and accelerating decoding.}
\label{fig:axon_effect}
\end{figure*}
Current fast parallel decoders may reveal several positions simultaneously if they are approximately conditionally independent. However, ``safe'' combinations don't always provide enough context for future steps, leading to states where the remaining positions are still too uncertain to decode safely. 
\method{} targets this limitation by augmenting a given decoder with a two stage strategy. First, \method{} computes a lightweight condition to determine if the base decoder needs additional contextual support. Second, if the condition is satisfied (i.e. the gate activates), then \method{} exposes one or more \textit{anchors}---which are masked positions that are both confident, and receive strong attention from uncertain positions. By revealing these anchors, \method{} exposes the necessary information required to reduce the uncertainty of the masked tokens. Figure~\ref{fig:axon_effect} illustrates this process.


\paragraph{Anchors.} 
Ideally, an anchor should be a position that reduces the uncertainty of several masked positions once revealed. Given a masked position $i$, the improvement of revealing position $j$ is ideally measured by the quantity $
    \Delta_i(j)
    =
    H(y_i\mid X,y^{(t)})
    -
    H(y_i\mid X,y^{(t)},y_j),
$
where $H$ denotes the entropy of the predictive distribution and $y_i,y_j$ are the tokens at positions $i$ and $j$. Optimizing $\Delta_i(j)$ directly is infeasible because the second term requires re-evaluating the model for each candidate anchor position. 

Inspired by previous attention-based decoding methods \cite{luo2026dawn, zhou2026attention}, we circumvent this limitation by using a \textit{lightweight} score based on the attention maps of the dLLM which are already available as a by-product of the denoising step. Explicitly, define
\begin{align}
    w_{ij}^{(t)}
    =
    A_{ij}^{(t)}
    \bigl(1-c_i^{(t)}\bigr)
    c_j^{(t)}, \label{eqn:score_def}
\end{align}
The confidence term $c_j^{(t)}$ selects for anchors that are confident and reliable, while the uncertainty term $(1-c_i^{(t)})$ emphasizes uncertain tokens that have the most need for additional context. Finally, the attention score $A_{ij}^{(t)}$ captures the fact that the uncertainty for a masked position $i$  has the most potential for improvement when it is strongly coupled with the anchor $j$. Indeed, the connection between attention and confidence improvement can be empirically verified (see Figure \ref{fig:confidence_attention2anchor}), further supported by the attention-dynamics and coverage analyses (see Appendix~\ref{sec:appendix_attention}), and made more rigorous for single-layer dLLMs (see Appendix \ref{app:confgain_attention_theory}).

\noindent
\textbf{Coverage objective.}
The anchor score above measures the usefulness of one candidate anchor for one uncertain position. However, when selecting multiple anchors at a denoising step (see Section~\ref{sec:experiments}), their value should not be treated independently. Two anchors may both appear useful, but if they support the same uncertain positions, revealing both gives little additional benefit. We therefore select anchors by maximizing a coverage objective under a cardinality constraint:
\begin{equation}
    \max_{S \subseteq V^{(t)},\, |S|\leq k}
    C^{(t)}(S)
    =
    \sum_{i\in U^{(t)}} \max_{j\in S} w_{ij}^{(t)} ,
\label{eq:coverage-objective}
\end{equation}
where 
$S$ is the selected anchor set, and
$k$ is the anchor budget. Each uncertain position receives credit only from its best selected anchor, encouraging selected anchors to cover different parts of the residual state. 

Proposition~\ref{prop:coverage-submodular} in Appendix~\ref{app:coverage-submodular-proof} proves that $C(S)$ (Eq.~\ref{eq:coverage-objective}) is \textit{monotone submodular}, giving diminishing returns when anchors cover overlapping uncertain positions. Formally, a set function $C$ is submodular if for every $X\subseteq Y$ and $j\notin Y$,
$$C(X\cup\{j\})-C(X)
\ge
C(Y\cup\{j\})-C(Y).$$
Submodular functions enjoy the property that under a cardinality budget $k$, simple greedy maximization gives a $(1-1/e)$ approximation guarantee~\citep{nemhauser1978analysis}. Overall, $C(S)$ encourages early (greedy) anchors to target the largest pockets of uncertainty, leading to total support rising quickly and then saturating. 

\begin{algorithm}[htb!]
\caption{\textcolor{codepurple}{\textbf{AXON}} decoding step at time $t$.}
\label{alg:sage}
\small
\vspace{1mm}
\textbf{Input:} Masked $M^{(t)}$, Uncertain $U^{(t)}$, Base $S^{(t)}_{\text{base}}$, Confidences $c^{(t)}$, Attention $A^{(t)}$, Block $B^{(t)}$, Gate weight $\alpha$, Budget $T$, Pace at step $t$ $r^{(t)}$, Coverage objective $C^{(t)}$      \\
\vspace{1mm}
{\color{gray}\makebox[\linewidth]{\dotfill}} \\
\vspace{1mm}
\textbf{Output:} Committed token set $S^{\star(t)}$
\vspace{1mm}
\hrule
\vspace{1mm}
\begin{algorithmic}[1]
\State \textcolor{codegreen}{\# compute decoding paces}
\State $r^{(t)} \gets |S^{(t)}_{\text{base}}| / |M^{(t)}|$
\State $\rho^{(t)} \gets 1 / (T - t)$
\vspace{1mm}
\State \textcolor{codegreen}{\# compute residual confidence and contextual support}
\State $\bar{c}_M^{(t)} \gets \frac{1}{|U^{(t)}|}\sum_{i\in U^{(t)}} c_i^{(t)}$
\State $g^{(t)} \gets \frac{1}{|U^{(t)}|}\sum_{i\in U^{(t)}} \frac{\sum_{j\in S^{(t)}_{\text{base}}} A_{ij}^{(t)}}{\sum_{j\in B^{(t)}} A_{ij}^{(t)} + \varepsilon}$ (Eq.~\ref{eq:g_t})
\vspace{1mm}
\State \textcolor{codegreen}{\# calculate deficits}
\State $d_{\text{pace}}^{(t)} \gets \big(\rho^{(t)} - r^{(t)}\big)\, \mathbb{I}\!\big[r^{(t)} \le r^{(t-1)}\,\big]$ (Eq.~\ref{eq:dpace})
\State $d_{\text{ctx}}^{(t)} \gets \max\!\big(r^{(t)},\, 1-\bar c_M^{(t)}\big) - g^{(t)}$ (Eq.~\ref{eq:dcov})
\vspace{1mm}
\State \textcolor{codegreen}{\# evaluate unified gate} (Eq.~\ref{eq:Gate})
\If{$\alpha\, \big(d_{\text{ctx}}^{(t)}\big)_+ + (1-\alpha)\, \big(d_{\text{pace}}^{(t)}\big)_+ > 0$}
    \State \textcolor{codegreen}{\# gate open: dynamic submodular selection}
    \State $\tau^{(t)} \gets C^{(t)}\big(S^{(t)}_{\text{base}}\big)$ (Eq.~\ref{eq:coverage-objective})
    \State $S \gets \emptyset$
    \While{$C^{(t)}(S) < \tau^{(t)}$ \textbf{and} $M^{(t)} \setminus S \neq \emptyset$}
        \State \textcolor{codegreen}{\# greedily add anchor with max marginal gain}
        \State $\Delta_j \gets C^{(t)}(S\cup\{j\}) - C^{(t)}(S)$
        \State $j^\star \gets \arg\max_{j \in M^{(t)} \setminus S} \Delta_j$
        \State $S \gets S \cup \{j^\star\}$
    \EndWhile
    \State $S^{\star(t)} \gets S$
\Else
    \State \textcolor{codegreen}{\# gate closed: commit base proposal}
    \State $S^{\star(t)} \gets S^{(t)}_{\text{base}}$
\EndIf
\vspace{1mm}
\State \Return $S^{\star(t)}$
\end{algorithmic}
\end{algorithm}

\noindent
\subsection{Gating mechanisms}
To effectively augment different base decoders with \method{}, we develop a mechanism to conditionally activate \method{} (i.e. a gate)  when the base decoder needs additional support. Our gate is computationally efficient as it depends solely on signals already computed during the forward pass. It detects two common failure modes of base decoders, namely (a) a ``pace deficit'', and (b) a ``context deficit''---which we detail in the immediate sequel.   

\paragraph{Pace deficit criterion.}
It is often desirable to enforce a fixed budget of diffusion steps, say $T \in \mathbb{N}$. At time step $t$ there are $|M^{(t)}|$ tokens that still need to be revealed and only $T-t$ steps remaining. An estimate of the \textit{fraction} of the remaining masked tokens that should be revealed at step $t$ to enforce this budget is,
$\rho^{(t)} = \frac{1}{T-t}$.
We interpret $\rho^{(t)}$ as the required \textit{pace}. Analogously, the base decoder's pace is simply the fraction of masked tokens revealed at step $t$, i.e. $r^{(t)} = \frac{|S^{(t)}_{\text{base}}|}{|M^{(t)}|}$. Thus, to detect a \textit{deficit} in the pace, we compute the following quantity, 
\begin{equation}
\label{eq:dpace}
d_{\text{pace}}^{(t)} = \big(\rho^{(t)} - r^{(t)}\big)\,
                         \mathbb{I}\!\big[r^{(t)} \le r^{(t-1)}\,\big].
\end{equation}
Here $\mathbb{I}[ \cdot ]$ denotes an indicator function. The term $\rho^{(t)} - r^{(t)}$ is a straightforward realization of the pace deficit, while the condition $r^{(t)} \le r^{(t-1)}$ allows \method{} to ignore the pace deficit when the base decoder's pace is still increasing, i.e. \method{} is optimistic that the pace deficit will be solved by the base decoder in subsequent steps. 
\begin{figure}
    \centering
    \includegraphics[width=0.9\linewidth]{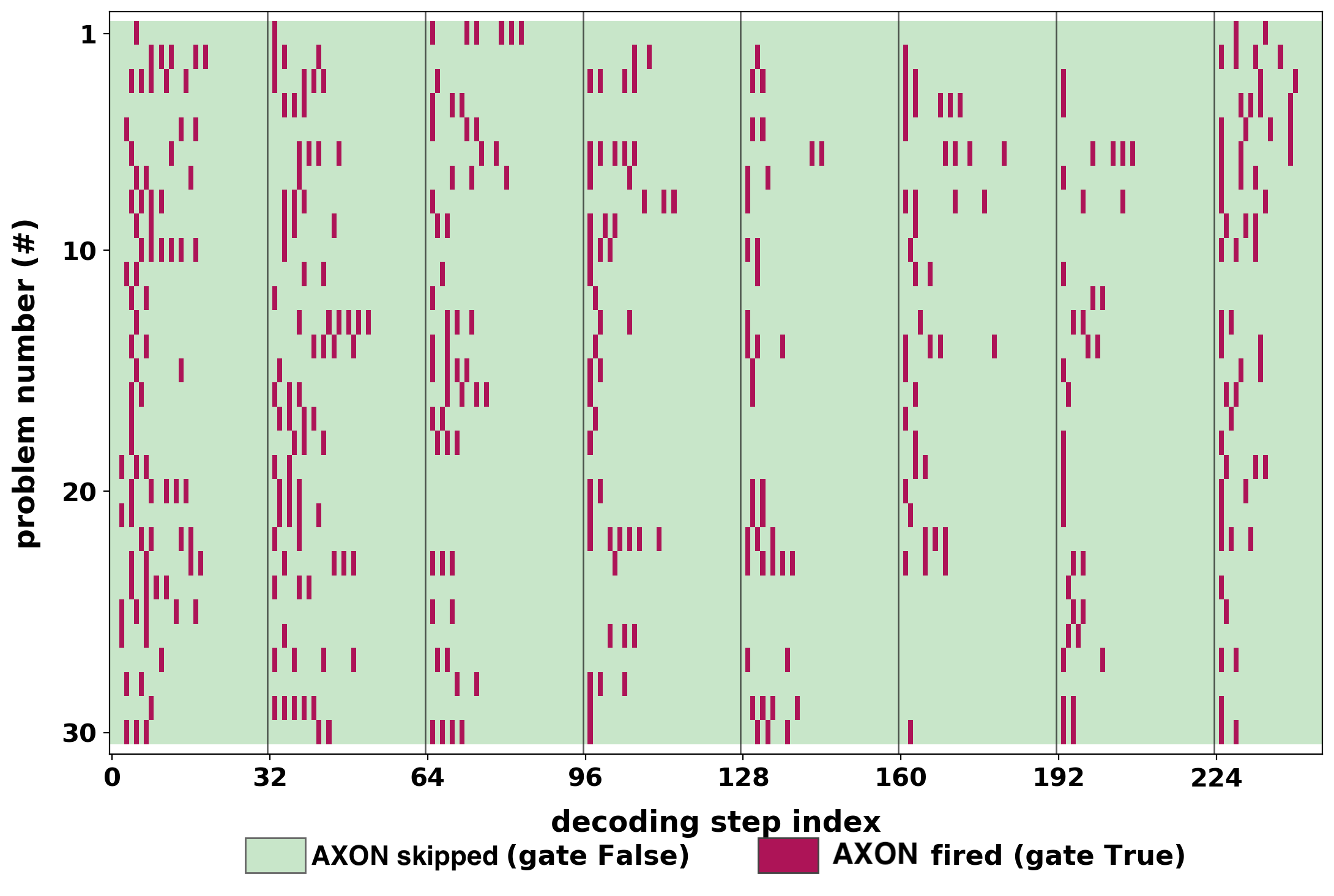}
    \caption{\textbf{Per-Step gate activity for LLaDA-1.5 + \method{} on GSM8K.} Rows are tasks in the dataset, columns are denoising steps (8 blocks of 32), red means that \method{} fired the gate and green means  it did not.}
    \label{fig:placeholder}
\end{figure}

\noindent
\textbf{Context deficit criterion.} One of the core goals of \method{} is to improve the contextual support provided by decoded tokens. Given the base decoder's proposal, we can compute a normalized measure of the contextual support as follows, 
\begin{equation}
\label{eq:g_t}
g^{(t)}
=
\frac{1}{|U^{(t)}|}
\sum_{i\in U^{(t)}}
\frac{
\sum_{j\in S^{(t)}_{\mathrm{base}}} A_{ij}^{(t)}
}{
\sum_{j\in B^{(t)}} A_{ij}^{(t)} + \varepsilon
}.
\end{equation}
Each term in the outer sum measures the support (in terms of attention) provided by $S^{(t)}_{\mathrm{base}}$ to position $i$, relative to the total support provided by all the tokens in the block. To detect low contextual support, we compare $g^{(t)}$ to a dynamic threshold to obtain our context deficit criterion, namely, 
\begin{equation}
\label{eq:dcov}
d_{\text{ctx}}^{(t)}  = \max\!\big(r^{(t)},\, u_M^{(t)}\big) - g^{(t)},
\end{equation}
where $r^{(t)}=\frac{|S^{(t)}_{\mathrm{base}}|}{|M^{(t)}|}$ as before, and $u_M^{(t)}= \frac{1}{|U^{(t)}|} \sum_{i\in U^{(t)}} (1-c_i^{(t)})$ is the mean uncertainty among uncertain tokens. The dynamic threshold captures two scenarios where we want to increase the contextual support further (by activating \method{}). Firstly, when $r^{(t)}$ is high, it means that the base decoder has committed a large number of positions, and so the contextual support should be proportionately high to avoid committing tokens that provide low support. Secondly, when $u_{M}^{(t)}$ is high, this means that the remaining masked tokens are highly uncertain, and so again, the contextual support should be increased proportionately. 
We refer the reader to Section~\ref{app:beta_ablation} in the Appendix for an ablation and additional analysis of alternative scalings of these two terms. In particular, one can introduce separate weights or normalization factors for the reveal ratio and uncertainty terms. However, our experiments show that keeping both signals on their natural scale, i.e., using the unweighted form in Eq.~\eqref{eq:dcov}, gives the best and most stable performance. This shows that the two quantities are already comparable enough for gating purposes.

\noindent
\textbf{Hybrid gate.} We combine both gates into an $\alpha$-weighted hybrid gate:
\begin{equation}\label{eq:Gate}
G_\alpha^{(t)} \;=\;
\mathbb{I}\!\Big[\,
\alpha\, \big(d_{\text{ctx}}^{(t)}\big)_+
+ (1-\alpha)\, \big(d_{\text{pace}}^{(t)}\big)_+
> 0
\,\Big],
\end{equation}
where $(x)_+ = \max(0, x)$ ensures that a negative deficit (surplus) in one criterion does not offset the other. When the gate is activated, \method{}'s anchor selection procedure kicks into action; otherwise, \method{} commits the base proposal. The effect of changing $\alpha$, which modulates the emphasis on the pace deficit and context criteria, is studied in Appendix ~\ref{apendix:alpha-ablation}. Activation patterns for $G_\alpha^{(t)}$ as a function of $t$ is visualized in Figure~\ref{fig:placeholder}. For additional analysis on gate firing patterns see Appendix~\ref{app:gate_distribution}.

%% file: latex/experiments.tex
\section{Experiments}
\label{sec:experiments}
We evaluate our framework against standard discrete diffusion benchmarks. We focus on 3 metrics: the quality-latency trade-off and the number of computational steps of our dynamic mechanism.
\input{tables/main_result}

\subsection{Setups}

\textbf{Models and datasets.}
We evaluate \method{} on three dLLM backbones spanning two model families: Dream-v0-Instruct-7B\footnote{https://huggingface.co/Dream-org/Dream-v0-Instruct-7B}~\citep{ye2025dream}, LLaDA-1.5\footnote{https://huggingface.co/GSAI-ML/LLaDA-1.5}~\citep{zhu2025llada}, and LLaDA-8B-Instruct\footnote{https://huggingface.co/GSAI-ML/LLaDA-8B-Instruct}~\citep{zhu2025llada}. This selection allows us to assess whether \method{} generalizes across architectures and training recipes. In all cases, we use the publicly released checkpoints as frozen backbones. We report results on four benchmarks covering reasoning and code generation: GSM8K~\citep{cobbe2021training} and Minerva-Math~\citep{hendrycks2021measuring} for mathematical problem solving, and HumanEval~\citep{chen2021evaluating} and MBPP~\citep{austin2021program} for Python program synthesis. All methods are evaluated under identical model weights, data splits, decoding budgets, and stopping criteria, so any observed differences can be attributed to the decoding strategy rather than the evaluation protocol. Additional implementation details can be found in Appendix~\ref{sec:implementation}. We additionally test generalization to a non-reasoning benchmark and a third backbone in Appendix~\ref{app:extra_datasets_backbones}.

\noindent
\textbf{Baselines.}
We compare against three recent training-free decoding strategies that cover the dominant design axes in the dLLM literature: confidence-based acceleration~\citep{wu2025fastdllm}, locality-aware scheduling (LocalLeap)~\citep{kong2025localleap}, and dependency-aware parallel commitment (DAWN)~\citep{luo2026dawn}. LocalLeap selects anchors by confidence and commits nearby positions within a local window, while DAWN schedules parallel commitments based on inter-token dependencies. These baselines are well aligned with our setting because they are training-free and operate through explicit anchor or commitment selection, allowing \method{} to be layered on top without modifying the underlying decoder.

\noindent
\textbf{Metrics.}
We report task accuracy, tokens per second (TPS), and number of function evaluations (NFE). Accuracy follows the standard evaluation protocol of each benchmark. TPS is the number of generated tokens divided by wall-clock decoding time, and NFE is the number of model forward passes required to complete generation. A desirable decoder improves accuracy or TPS while reducing NFE, or improves one metric without a comparable drop in the others.

\noindent

\textbf{Number of Anchors.}
\method{} is intended to provide minimal contextual support rather than to replace the base decoder's commit rule. We therefore evaluate a conservative default variant with a fixed budget of $k=1$ anchor per gated intervention. We also report \method{}\textsubscript{CVR}, an adaptive variant that greedily adds anchors until their coverage matches the contextual support induced by the base decoder's proposal, i.e., until $C^{(t)}(S_m)\ge C^{(t)}(S^{(t)}_{\mathrm{base}})$. See Appendix~\ref{app:anchor_ablation} for a study of the effect of the anchor budget.

\subsection{Results}
Table~\ref{tab:main} summarizes accuracy and latency across all baselines, comparing each base decoder with the fixed-budget \method{} variant and the adaptive \method{}\textsubscript{CVR} variant. Across models, tasks, and decoders, \method{} improves accuracy or TPS in nearly every setting while consistently reducing NFE.

Notably, several settings show simultaneous gains in accuracy and throughput. On LLaDA-8B-Instruct with LocalLeap, \method{} improves HumanEval accuracy by $3.05$ points while increasing TPS, and \method{}\textsubscript{CVR} further improves throughput while remaining $2.43$ accuracy points above the baseline. Similar trends appear on LLaDA-$1.5$. On mathematical reasoning, \method{} improves both accuracy and TPS on GSM8K and Minerva-Math with LLaDA-$1.5$ and DAWN, while reducing NFE. On Dream-v0-Instruct-7B, DAWN+\method{} improves GSM8K accuracy by $1.67$ points with higher TPS, and improves both accuracy and throughput on MBPP with lower NFE. Additional per-decoder analysis is provided in Section~\ref{sec:app_affect_on_decoders}.


\subsection{Ablations}
\label{sec:ablations}
A natural alternative to our objective~\eqref{eq:coverage-objective}
is additive coverage,
${\mathrm{GC}}(S)=\frac{1}{|U^{(t)}|}\sum_{i\in U^{(t)}}\sum_{j\in S} w_{ij},$
which is also a submodular function. The key difference is that objective \eqref{eq:coverage-objective} (also known as \textit{facility-location}) assigns credit from each uncertain position only to its best selected anchor, while additive coverage rewards all anchors which have positive weight. As a result, optimizing $\mathrm{GC}(S)$ can repeatedly select anchors that support the same uncertain positions, whereas facility-location naturally favors diverse and non-redundant reveals through diminishing returns.
Table~\ref{tab:fl-vs-gc} compares the two objectives. Facility-location consistently achieves lower NFE while maintaining comparable accuracy, indicating that explicitly encouraging diverse anchor coverage improves decoding efficiency. In Appendix~\ref{app:submod-ablation}, We further tested whether adding an explicit redundancy penalty improves selection quality. In practice, the effect was negligible, suggesting that facility-location already discourages redundant anchors sufficiently through its saturating coverage structure. We therefore use the monotone facility-location objective throughout the paper. In addition to the anchor selection objective, the performance of \method{} relies on the unified gate weight $\alpha$, which balances the framework's sensitivity between context and pace deficits. Based on our evaluations, we configure \method{} with $\alpha=0.2$ for the LLaDA and $\alpha=0.9$ for Dream-Instruct across all main-text experiments; a comprehensive ablation of these gating dynamics and their impact on throughput and accuracy is provided in Appendix~\ref{apendix:alpha-ablation}.
\begin{table}[htb!]
\centering
\scriptsize
\setlength{\tabcolsep}{3.0pt}
\resizebox{0.9\linewidth}{!}{%
\begin{tabular}{ll ccc ccc}
\toprule
& & \multicolumn{3}{c}{$C(S)$} & \multicolumn{3}{c}{$\text{GC}(S)$} \\
\cmidrule(lr){3-5} \cmidrule(lr){6-8}
\textbf{Model} & \textbf{Task} & Acc & TPS & NFE & Acc & TPS & NFE \\
\midrule
Dream-7B   & MBPP      & 55.00 & 29.76 & \textbf{41.81} & 54.60 & 29.98 & 42.25 \\
LLaDA-1.5  & MBPP      & 35.20 & 21.23 & \textbf{28.30} & 35.00 & 20.69 & 29.33 \\
LLaDA-8B   & MBPP      & 30.40 & 32.01 & \textbf{42.40} & 29.80 & 31.69 & 43.46 \\
LLaDA-1.5  & HumanEval & 39.02 & 28.73 & \textbf{59.49} & 41.46 & 28.41 & 60.49 \\
LLaDA-8B   & HumanEval & 39.98 & 90.64 & \textbf{52.43} & 39.98 & 88.27 & 54.22 \\
\midrule
\multicolumn{2}{l}{Mean across cells}
  & \textit{33.14} & \textit{\textbf{40.47}} & \textit{\textbf{44.89}}
  & \textit{\textbf{33.39}} & \textit{39.81} & \textit{45.95} \\
\bottomrule
\end{tabular}
}
\caption{\textbf{Ablating coverage form.} Results comparing our objective, monotone facility-location ($C(S)$) against the graph-cut (GC) across tasks and models.}
\label{tab:fl-vs-gc}
\end{table}

\section{Conclusion and Future Work}
We introduced \method{}, a training-free module for improving parallel decoding in diffusion language models. \method{} monitors the decoding state and intervenes only when the current revealed tokens provide insufficient support to the remaining masked positions. When needed, it reveals a small set of influential and non-redundant anchors that provide targeted context for subsequent denoising steps.

Across reasoning and code-generation benchmarks, multiple dLLM backbones, and different decoding strategies, \method{} improves the quality-latency trade-off without changing the backbone or base decoder. Our ablations show that state monitoring and coverage-based anchor selection matter, suggesting that efficient decoding should go beyond commit rules and construct useful context.

More broadly, our work sheds light on the importance of state monitoring and anchor selection in future parallel decoding, whether through gating mechanisms or self-monitoring strategies detecting when contextual support is needed.

\section{Limitations}
\method{} improves existing parallel decoders through gated anchor interventions, but it has several limitations. First, it relies on attention and confidence signals from the frozen dLLM. These signals are lightweight and available during decoding, but they are only proxies for token influence and uncertainty. When confidence is poorly calibrated or attention does not reflect useful dependencies, anchor selection may provide limited benefit.

Second, \method{} adds scoring and selection overhead. In our setting this cost is controlled because the gate fires only when needed and anchor selection is performed over a limited candidate set. For longer sequences or larger candidate pools, more approximate selection may be required.

Finally, \method{} still has a quality-latency trade-off through the gate, anchor budget, and base decoder configuration. Our experiments show favorable operating points across multiple models, benchmarks, and decoder families, but the best configuration may vary by task and decoding budget. We focus on text-only reasoning and code-generation tasks; long-context, dialogue, multilingual, and multimodal settings remain future work.

\section{AI Use Statement}

In this work, we used generative AI tools for text polishing and for writing code.
Additionally, we used generative AI tools for part of the ideation
and to assist with results analysis.

We have reviewed all AI-assisted work. AI-generated code was verified and stress
tested by three of the authors. AI-polished text was read and edited
by an author to confirm that it preserves the intended technical meaning and
introduces no claim beyond what our experiments support.

%% file: tables/main_result.tex
\definecolor{lightgreen}{HTML}{E2F0D9} 

\begin{table*}[htb!]
\centering
\footnotesize
\setlength{\tabcolsep}{3pt}
\resizebox{0.9\linewidth}{!}{%
\begin{tabular}{l|rrr|rrr|rrr|rrr}
\hline
& \multicolumn{3}{c|}{\underline{HumanEval}} & \multicolumn{3}{c|}{\underline{MBPP}} & \multicolumn{3}{c|}{\underline{GSM8K}} & \multicolumn{3}{c}{\underline{Minerva-Math}} \\
Method & Acc ($\uparrow$) & TPS($\uparrow$) & NFE($\downarrow$) & Acc ($\uparrow$) & TPS($\uparrow$) & NFE($\downarrow$) & Acc ($\uparrow$) & TPS($\uparrow$) & NFE($\downarrow$) & Acc ($\uparrow$) & TPS($\uparrow$) & NFE($\downarrow$) \\
\hline
\multicolumn{13}{c}{\textit{LLaDA-1.5}}\\
\hline
Original & 43.90 & 6.04 & 256.0 & 38.80 & 2.39 & 256.0 & 80.36 & 6.70 & 256.0 & 33.36 & 8.35 & 256.0 \\
\hline
Confidence & 43.29 & 15.61 & 100.6 & 38.00 & 14.11 & 42.5 & 80.44 & 22.95 & 74.8 & \textbf{33.38} & 22.31 & 95.4 \\
\quad + \method{} & \colorbox{lightgreen}{43.90} & \colorbox{lightgreen}{18.53} & \colorbox{lightgreen}{93.6} & \colorbox{lightgreen}{\textbf{39.20}} & \colorbox{lightgreen}{14.39} & \colorbox{lightgreen}{42.2} & \colorbox{lightgreen}{\textbf{81.50}} & 22.57 & \colorbox{lightgreen}{72.1} & 33.24 & \colorbox{lightgreen}{22.61} & \colorbox{lightgreen}{92.7} \\
\quad + \method{}\textsubscript{CVR} & 43.29 & \colorbox{lightgreen}{19.74} & \colorbox{lightgreen}{91.0} & \colorbox{lightgreen}{38.20} & \colorbox{lightgreen}{14.52} & \colorbox{lightgreen}{42.1} & \colorbox{lightgreen}{81.27} & \colorbox{lightgreen}{23.30} & \colorbox{lightgreen}{70.3} & 33.02 & \colorbox{lightgreen}{23.06} & \colorbox{lightgreen}{91.1} \\
\hdashline
LocalLeap & 41.46 & 19.83 & 75.2 & 38.80 & 18.50 & 32.5 & 80.06 & 29.46 & 57.3 & 32.40 & 27.64 & 75.7 \\
\quad + \method{} & \colorbox{lightgreen}{\textbf{44.51}} & \colorbox{lightgreen}{20.42} & 77.6 & 38.20 & 16.77 & 35.3 & \colorbox{lightgreen}{80.74} & 26.74 & 60.2 & \colorbox{lightgreen}{32.88} & 25.86 & 79.7 \\
\quad + \method{}\textsubscript{CVR} & \colorbox{lightgreen}{\textbf{44.51}} & \colorbox{lightgreen}{22.91} & \colorbox{lightgreen}{73.0} & 37.60 & 17.13 & 34.6 & \colorbox{lightgreen}{81.12} & 27.64 & 58.1 & 32.34 & 26.47 & 78.1 \\
\hdashline
DAWN & 42.07 & 22.96 & 77.3 & 37.80 & 19.37 & 30.7 & 80.82 & 30.55 & 54.2 & 31.80 & 28.63 & 72.9 \\
\quad + \method{} & \colorbox{lightgreen}{42.68} & \colorbox{lightgreen}{26.96} & \colorbox{lightgreen}{62.9} & 37.20 & \colorbox{lightgreen}{20.97} & \colorbox{lightgreen}{29.3} & \colorbox{lightgreen}{81.20} & \colorbox{lightgreen}{33.02} & \colorbox{lightgreen}{49.5} & \colorbox{lightgreen}{32.00} & \colorbox{lightgreen}{30.64} & \colorbox{lightgreen}{68.0} \\
\quad + \method{}\textsubscript{CVR} & 39.02 & \colorbox{lightgreen}{\textbf{29.11}} & \colorbox{lightgreen}{\textbf{59.6}} & 37.20 & \colorbox{lightgreen}{\textbf{21.49}} & \colorbox{lightgreen}{\textbf{28.3}} & 80.52 & \colorbox{lightgreen}{\textbf{34.41}} & \colorbox{lightgreen}{\textbf{47.6}} & 31.30 & \colorbox{lightgreen}{\textbf{31.60}} & \colorbox{lightgreen}{\textbf{66.0}} \\
\hline\hline
\multicolumn{13}{c}{\textit{LLaDA-8B-Instruct}}\\
\hline
Original & 40.24 & 18.09 & 256.0 & 29.40 & 6.48 & 256.0 & 77.94 & 7.14 & 256.0 & \textbf{33.20} & 9.51 & 256.0 \\
\hline
Confidence & 40.85 & 58.96 & 77.9 & 29.60 & 23.69 & 69.8 & 78.39 & 23.48 & 77.7 & 32.98 & 25.14 & 96.9 \\
\quad + \method{} & 40.55 & \colorbox{lightgreen}{65.45} & \colorbox{lightgreen}{74.6} & \colorbox{lightgreen}{32.00} & 23.51 & \colorbox{lightgreen}{62.2} & \colorbox{lightgreen}{\textbf{79.30}} & 23.39 & \colorbox{lightgreen}{74.7} & 32.76 & \colorbox{lightgreen}{25.75} & \colorbox{lightgreen}{92.7} \\
\quad + \method{}\textsubscript{CVR} & 40.55 & \colorbox{lightgreen}{66.23} & \colorbox{lightgreen}{73.6} & \colorbox{lightgreen}{31.40} & 23.69 & \colorbox{lightgreen}{61.6} & \colorbox{lightgreen}{79.15} & \colorbox{lightgreen}{24.16} & \colorbox{lightgreen}{72.2} & 32.30 & \colorbox{lightgreen}{26.23} & \colorbox{lightgreen}{91.0} \\
\hdashline
LocalLeap & 39.63 & 71.20 & 61.4 & 29.40 & 29.84 & 55.0 & 77.33 & 30.55 & 58.9 & 32.42 & 31.61 & 75.6 \\
\quad + \method{} & \colorbox{lightgreen}{\textbf{42.68}} & \colorbox{lightgreen}{76.66} & 63.1 & \colorbox{lightgreen}{31.40} & 28.05 & \colorbox{lightgreen}{51.5} & \colorbox{lightgreen}{77.48} & 27.97 & 62.1 & \colorbox{lightgreen}{32.62} & 29.75 & 79.1 \\
\quad + \method{}\textsubscript{CVR} & \colorbox{lightgreen}{42.06} & \colorbox{lightgreen}{79.39} & \colorbox{lightgreen}{60.2} & \colorbox{lightgreen}{30.20} & 28.61 & \colorbox{lightgreen}{50.6} & 75.89 & 28.90 & 59.9 & 32.16 & 30.52 & 77.2 \\
\hdashline
DAWN & 40.24 & 80.26 & 63.2 & 30.80 & 31.19 & 50.0 & 77.94 & 31.95 & 55.3 & 32.36 & 32.83 & 72.5 \\
\quad + \method{} & 39.98 & \colorbox{lightgreen}{92.00} & \colorbox{lightgreen}{53.2} & \colorbox{lightgreen}{\textbf{32.40}} & \colorbox{lightgreen}{31.69} & \colorbox{lightgreen}{44.4} & 77.41 & \colorbox{lightgreen}{34.33} & \colorbox{lightgreen}{50.6} & 31.84 & \colorbox{lightgreen}{35.36} & \colorbox{lightgreen}{67.0} \\
\quad + \method{}\textsubscript{CVR} & 39.58 & \colorbox{lightgreen}{\textbf{92.22}} & \colorbox{lightgreen}{\textbf{52.4}} & 30.60 & \colorbox{lightgreen}{\textbf{32.35}} & \colorbox{lightgreen}{\textbf{42.4}} & 76.57 & \colorbox{lightgreen}{\textbf{35.89}} & \colorbox{lightgreen}{\textbf{48.2}} & 30.84 & \colorbox{lightgreen}{\textbf{36.57}} & \colorbox{lightgreen}{\textbf{64.7}} \\
\hline\hline
\multicolumn{13}{c}{\textit{Dream-v0-Instruct-7B}}\\
\hline
Original & 53.66 & 16.62 & 256.0 & 54.20 & 5.42 & 256.0 & \textbf{76.35} & 4.85 & 256.0 & 37.80 & 11.50 & 256.0 \\
\hline
Confidence & \textbf{57.32} & 38.30 & 90.4 & 55.20 & 26.38 & 62.7 & 75.20 & 19.23 & 62.3 & 38.24 & 24.66 & 114.8 \\
\quad + \method{} & 57.30 & \colorbox{lightgreen}{41.68} & \colorbox{lightgreen}{75.4} & \colorbox{lightgreen}{55.40} & 25.19 & \colorbox{lightgreen}{56.0} & \colorbox{lightgreen}{76.12} & 18.53 & \colorbox{lightgreen}{61.6} & \colorbox{lightgreen}{\textbf{38.50}} & 24.04 & \colorbox{lightgreen}{112.1} \\
\quad + \method{}\textsubscript{CVR} & 57.30 & \colorbox{lightgreen}{42.90} & \colorbox{lightgreen}{70.2} & 55.20 & 24.98 & \colorbox{lightgreen}{55.1} & \colorbox{lightgreen}{75.82} & 18.44 & \colorbox{lightgreen}{61.2} & 37.74 & 24.20 & \colorbox{lightgreen}{110.0} \\
\hdashline
LocalLeap & 54.27 & 42.79 & 76.1 & 54.60 & 28.69 & 44.7 & 73.24 & 22.05 & 54.2 & 38.10 & 28.22 & 97.7 \\
\quad + \method{} & \colorbox{lightgreen}{54.90} & \colorbox{lightgreen}{47.60} & \colorbox{lightgreen}{65.1} & \colorbox{lightgreen}{55.60} & 28.47 & \colorbox{lightgreen}{41.4} & \colorbox{lightgreen}{75.06} & 21.97 & \colorbox{lightgreen}{51.5} & 37.90 & \colorbox{lightgreen}{28.38} & \colorbox{lightgreen}{91.2} \\
\quad + \method{}\textsubscript{CVR} & 54.27 & \colorbox{lightgreen}{47.80} & \colorbox{lightgreen}{70.0} & \colorbox{lightgreen}{55.00} & 28.25 & \colorbox{lightgreen}{\textbf{40.6}} & \colorbox{lightgreen}{74.91} & 22.04 & \colorbox{lightgreen}{\textbf{51.1}} & 37.42 & \colorbox{lightgreen}{28.32} & \colorbox{lightgreen}{\textbf{89.5}} \\
\hdashline
DAWN & 54.88 & 44.77 & 76.2 & 55.40 & 29.76 & 42.8 & 73.16 & 22.20 & 52.08 & 38.22 & 28.65 & 94.5 \\
\quad + \method{} & \colorbox{lightgreen}{55.60} & \colorbox{lightgreen}{49.70} & \colorbox{lightgreen}{\textbf{64.9}} & \colorbox{lightgreen}{\textbf{55.80}} & \colorbox{lightgreen}{\textbf{30.09}} & \colorbox{lightgreen}{42.6} & \colorbox{lightgreen}{74.83} & \colorbox{lightgreen}{22.59} & \colorbox{lightgreen}{51.3} & \colorbox{lightgreen}{38.28} & \colorbox{lightgreen}{\textbf{28.89}} & \colorbox{lightgreen}{93.0} \\
\quad + \method{}\textsubscript{CVR} & 54.80 & \colorbox{lightgreen}{\textbf{50.10}} & \colorbox{lightgreen}{68.8} & 55.00 & \colorbox{lightgreen}{29.89} & \colorbox{lightgreen}{41.8} & \colorbox{lightgreen}{74.53} & \colorbox{lightgreen}{\textbf{22.67}} & \colorbox{lightgreen}{51.7} & 37.76 & \colorbox{lightgreen}{28.86} & \colorbox{lightgreen}{91.2} \\
\hline
\end{tabular}
}
\caption{\textbf{Main quantitative evaluations.} We evaluate across selection rules our submodular anchor framework. We report task Accuracy (\%), Throughput (TPS), and Number of Function Evaluations (NFE). \colorbox{lightgreen}{Green} cell background shading next to a \method{} value indicates an improvement over the parent baseline decoder ($\uparrow$ for Acc and TPS, $\downarrow$ for NFE). Best values overall per backbone and per dataset metric are in \textbf{bold}.}
\label{tab:main}
\end{table*}

%% file: latex/ack.tex
\newcommand{\acksection}{\section*{Acknowledgments and Disclosure of Funding}}
\NewEnviron{ack}{%
  \acksection
  \BODY
}
\begin{ack}
The authors thank the Stuttgart Center for Simulation Science (SimTech) and the International Max Planck Research School for Intelligent Systems (IMPRS-IS) for supporting Lluís, Aneesh, Tanja, and Loay. This research was funded by the Ministry of Science, Research and the Arts Baden-Wuerttemberg through the Artificial Intelligence Software Academy (AISA). We also acknowledge the support of the Stuttgart Center for Simulation Science (SimTech). Tanja was supported by the Deutsche Forschungsgemeinschaft (DFG, German Research Foundation) under Germany's Excellence Strategy -- EXC 2120/1 -- 390831618. Aneesh was supported and funded by the Deutsche Forschungsgemeinschaft (DFG, German Research Foundation) under Germany's Excellence Strategy -- EXC 2075 -- 390740016. L.\ Mualem was supported by a postdoctoral scholarship from the Planning and Budgeting Committee (PBC) of the Council for Higher Education in Israel. L.\ Mualem gratefully acknowledge the computing time provided on the high-performance computer HoreKa by the National High-Performance Computing Center at KIT (NHR@KIT). This center is jointly supported by the Federal Ministry of Education and Research and the Ministry of Science, Research and the Arts of Baden-Württemberg, as part of the National High-Performance Computing (NHR) joint funding program (https://www.nhr-verein.de/en/our-partners). HoreKa is partly funded by the German Research Foundation (DFG).
A.\ Maalouf acknowledges support from the Neubauer Family Foundation and from the MAOF Fellowship of the Council for Higher Education.

\end{ack}2

%% file: latex/appendix.tex
    \appendix
\section{Additional Related Work}
\label{sec:app_related}
\paragraph{Adaptive and revocable commitment.} 
A related line changes how aggressive commitments are organized or
revised over time. WINO~\citep{hong2025wino} re-masks suspicious commits to make aggressive drafts less risky, while RDD~\citep{wang2026rdd} rolls back previous blocks when the current block stalls. DCD~\citep{shu2026dcd} replaces rigid block boundaries with a confidence-aware sliding window. SlowFast~\citep{wei2025accelerating}
alternates between an exploratory stage that identifies stable
high-confidence spans and an accelerated stage that decodes them in
parallel.
Each controls the risk, consequence, or timing of a commit. \method{}
does not replace these mechanisms. It decides which additional reveal is
worth making once the decoder's safety rule has been applied.

\paragraph{Model-side and systems-level acceleration.} 
d3LLM~\citep{qian2026d3llm} accelerates dLLMs by retraining the model so it can expose early-decodable tokens sooner. This places d3LLM on the model-training side of the acceleration landscape, whereas AXON keeps the backbone frozen and modifies only the inference-time commit rule. 

An orthogonal line reduces wall-clock cost by changing how the model is
evaluated or verified, not which positions are committed. Key-value caching for
dLLMs~\citep{wu2025fastdllm} avoids repeated computation across
denoising steps, and speculative systems such as
Spiffy~\citep{agrawal2025spiffy} accepts many drafted reveals in parallel
under a lossless distribution-preserving test. This is orthogonal to the
commit-selection problem we address: systems methods make individual
steps cheaper, while \method{} changes the commit mask when the host
decoder's own rule leaves the residual under-covered. The two compose, so we treat system acceleration as complementary to anchor selection.

\paragraph{Submodular optimization.}
Submodular functions model diminishing returns, where the marginal gain of adding a new element decreases as the selected set grows. In our setting, once an uncertain position is already well supported by one selected anchor, adding another anchor that supports the same position should provide less additional value. This naturally motivates a facility-location objective, where each uncertain position receives credit only from its best selected anchor. Beyond our setting, submodular objectives are widely used in machine learning for selecting representative and non-redundant subsets, including data summarization~\cite{chen2024guided,mualem2024submodular}, active learning~\cite{chen2013near}, feature selection~\cite{tukan2024practical}, model compression~\cite{el2022data}, and sensor placement~\citep{zhou2022risk,tukan2023orbslam3}. Greedy optimization of monotone submodular functions further admits the standard $(1-1/e)$ approximation guarantee under cardinality constraints~\citep{nemhauser1978analysis}.
\section{Hardware and Implementation Details.}
\label{sec:implementation}
All experiments were implemented within the DAWN~\footnote{https://github.com/lizhuo-luo/DAWN} diffusion decoding codebase, reusing its model weights, tokenizer settings, decoding budgets, stopping criteria, and block-sampling protocol. Throughput is measured end-to-end and includes the overhead introduced by \method{}'s commit-selection step during decoding. Experiments are conducted on a cluster of 8$\times$NVIDIA A100 80GB GPUs. 
We set $\beta_r=\beta_u=1$ since all the involved quantities are normalized in the interval $[0,1]$.

\begin{figure*}[t]
    \centering
    \includegraphics[width=1.0\linewidth]{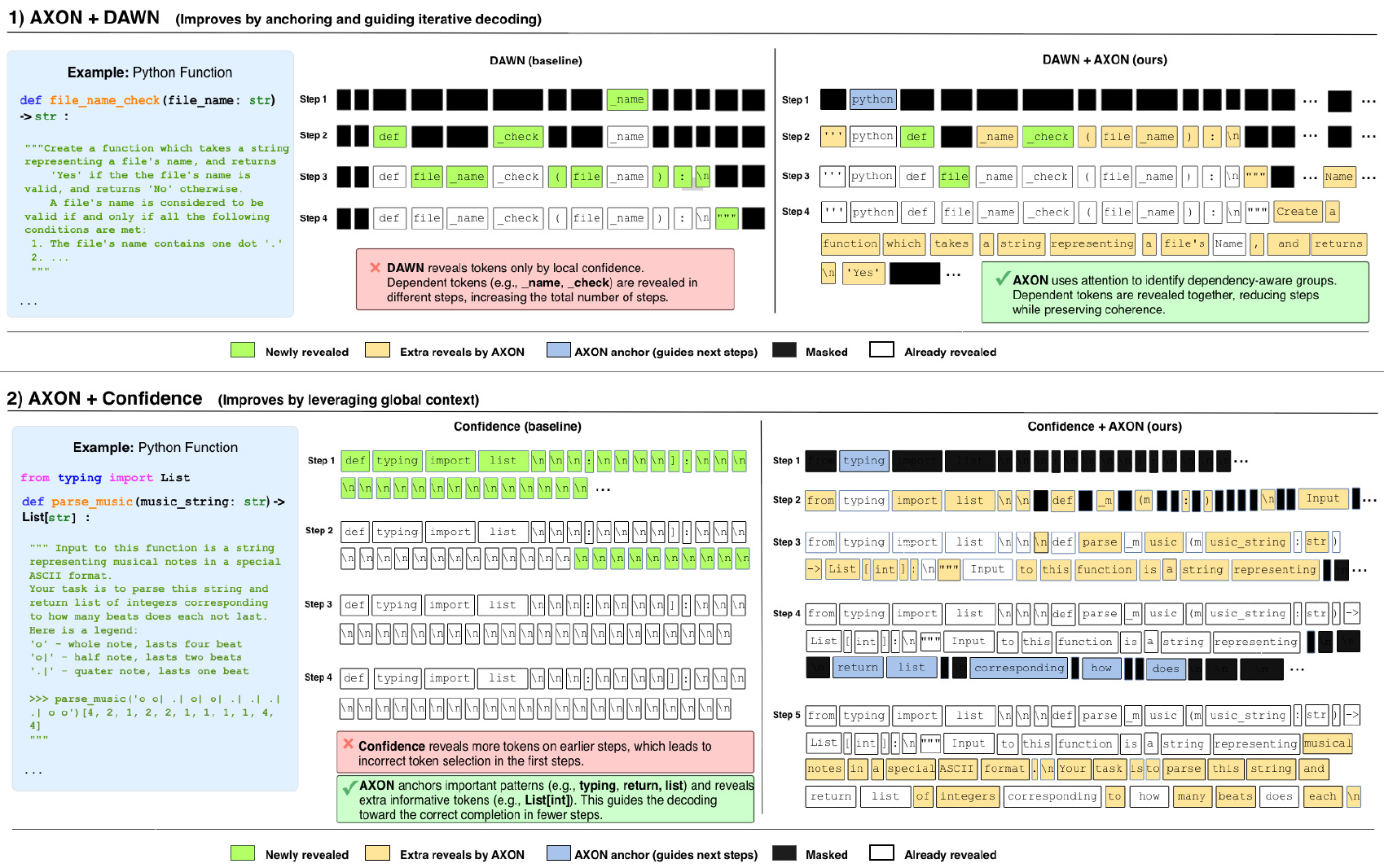}

\caption{\textbf{Comparison of token reveals across decoding methods.} (1) shows the trajectory of DAWN, (2) tracks the Confidence decoder. In both sets, the base decoder sits on the left figure and our anchor-guided \method{} variant sits to the right. \textcolor{green!60!black}{Green}: tokens newly unmasked at the current step. \textcolor{blue!70!gray}{Blue}: anchor tokens selected by \method{} at step $i$ to guide reveals at steps $>i$.
\textcolor{orange}{Orange}: tokens unmasked because of the anchors selected by \method{}.  {blocks}: remaining masked positions. \method{}'s anchor mechanism unlocks semantically coherent token groups significantly earlier, reducing total NFE while enhancing context coverage.}
    \label{fig:coverage_trajectory}
\end{figure*}

\section{Attention Dynamics and Coverage Analysis}
\label{sec:appendix_attention}

To better understand the mechanistic impact of \method{} on the diffusion decoding process, we probe the internal attention dynamics of the model during generation. Specifically, we analyze how attention patterns evolve across inference steps, how the revelation of structural anchors shifts these patterns, and how \method{} accelerates the accumulation of contextual coverage. 
In Section~\ref{sec:app_qualitative_token_dynamics} evaluates LLaDA-8B Instruct on the HumanEval dataset, comparing the base decoders (Confidence, LocalLeap, and DAWN) with and without our proposed method, \method{}. In Section~\ref{sec:app_Mechanistic Impact of Anchor Commits}, we evaluate LLaDA-1.5 on the MBPP dataset, comparing the base DAWN decoder against DAWN equipped with \method{}. Similarly, Section~\ref{sec:app_NFE_improve_analysis} evaluates LLaDA-1.5 on the HumanEval dataset, comparing the base DAWN decoder against DAWN with \method{}.

\subsection{Qualitative Analysis of Token Dynamics}\label{sec:app_qualitative_token_dynamics}
Figure~\ref{fig:coverage_trajectory} visualizes the step-by-step decoding trajectories to illustrate how~\method{} optimizes the parallel reveal mechanism compared to standard decoders. In the baseline configurations (left panels), both DAWN and Confidence decoders exhibit fragmented reveals. DAWN gets bottlenecked by local token dependencies, forcing sequential steps to resolve closely coupled syntax like \_name and \_check. Confidence decoding (bottom left) suffers from early global scattering, unmasking generic syntax elements ($\texttt{\textbackslash n}$) prematurely without semantic coordination.
In contrast, when using ~\method{} on top of the decoders (right panels), it intercepts these under-supported states. By injecting critical structural anchors (\textcolor{blue!70!gray}{blue tokens} like python or typing), ~\method{} leverages global attention paths to safely unmask entire dependent blocks simultaneously (\textcolor{orange}{orange tokens}). This effectively resolves hard dependency cores blocks ahead of schedule, thus minimizing the total NFE required to achieve coherent code generation.

\subsection{Mechanistic Impact of Anchor Commits}\label{sec:app_Mechanistic Impact of Anchor Commits}

\begin{figure}[t]

    \centering

    \includegraphics[width=\linewidth]{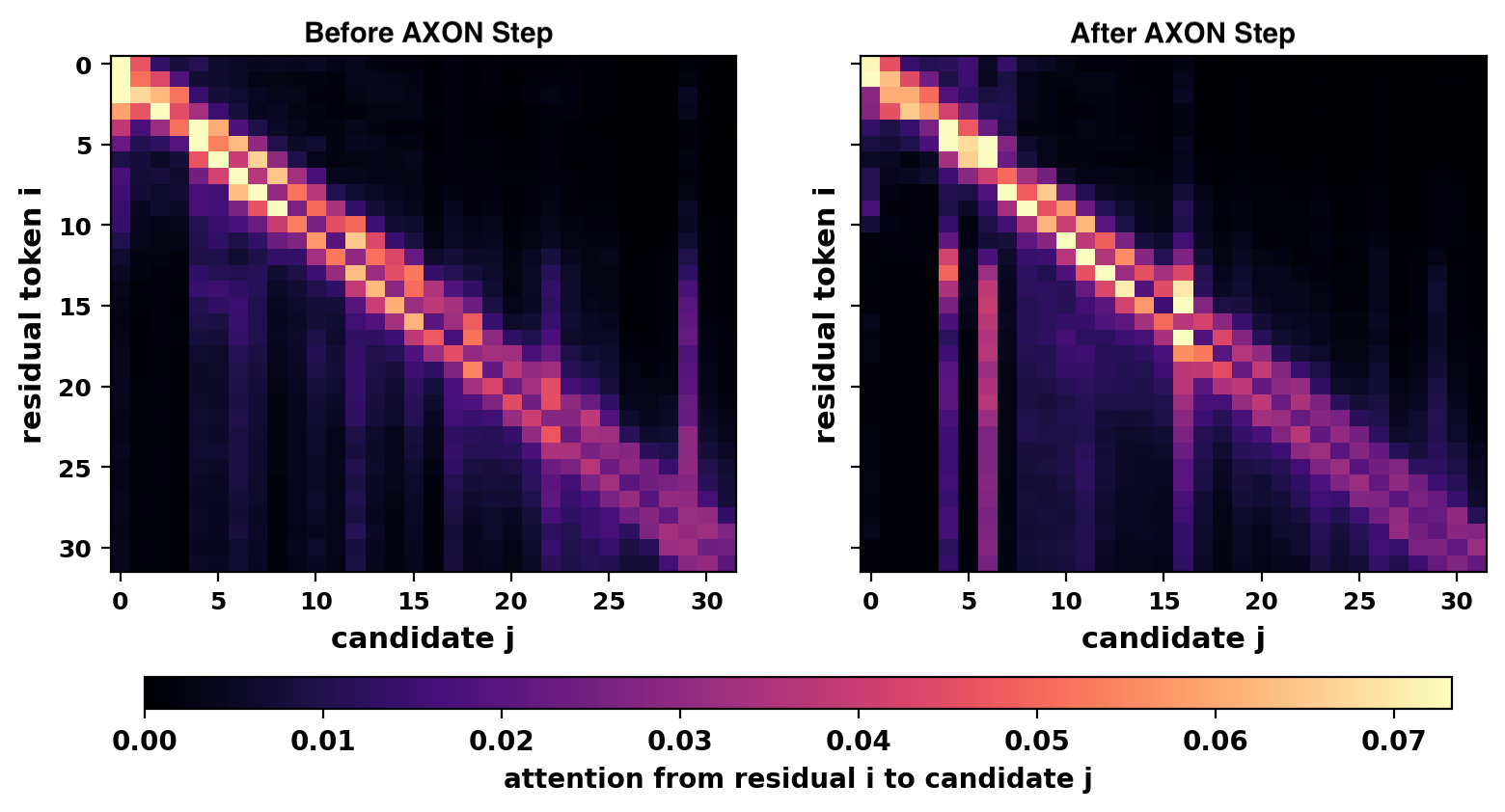}

    \caption{\textbf{Attention routing before and after a \method{} intervention.} The $32 \times 32$ attention slice for a single block (Block 0, Step 0 to Step 1) of an MBPP problem. \textbf{Left:} The step immediately before \method{} fires. \textbf{Right:} The step immediately after \method{} anchors are committed. The emergence of bright vertical streaks, such as columns 4, 6, and 16, shows newly revealed anchor tokens absorbing attention mass from the remaining masked residuals.}

    \label{fig:attn_before_after}

\end{figure}

When \method{} fires, it explicitly selects masked tokens that are highly attended to by the remaining uncertain positions. Figure~\ref{fig:attn_before_after} illustrates the immediate structural impact of revealing these anchors.

We visualize the raw $32 \times 32$ attention slice (where rows indicate the attending position and columns indicate the attended position) just before and immediately after \method{} commits its selected anchors. In the pre-commit state, attention is largely dominated by local, diagonal structures. However, in the post-commit state, distinct vertical streaks appear corresponding exactly to the positions \method{} just unmasked. Once these structural anchors are converted from \texttt{[MASK]} to real tokens, they become highly informative landmarks. The remaining residual tokens rapidly route their attention mass to these new visible contexts, confirming that the chosen anchors successfully absorb the unresolved dependency structure of the block.

\subsection{Token Unmasking Dynamics}\label{sec:app_NFE_improve_analysis}
\begin{figure*}[t]
  \centering
  
  \begin{subfigure}{0.48\linewidth}
    \centering
    \includegraphics[width=\linewidth]{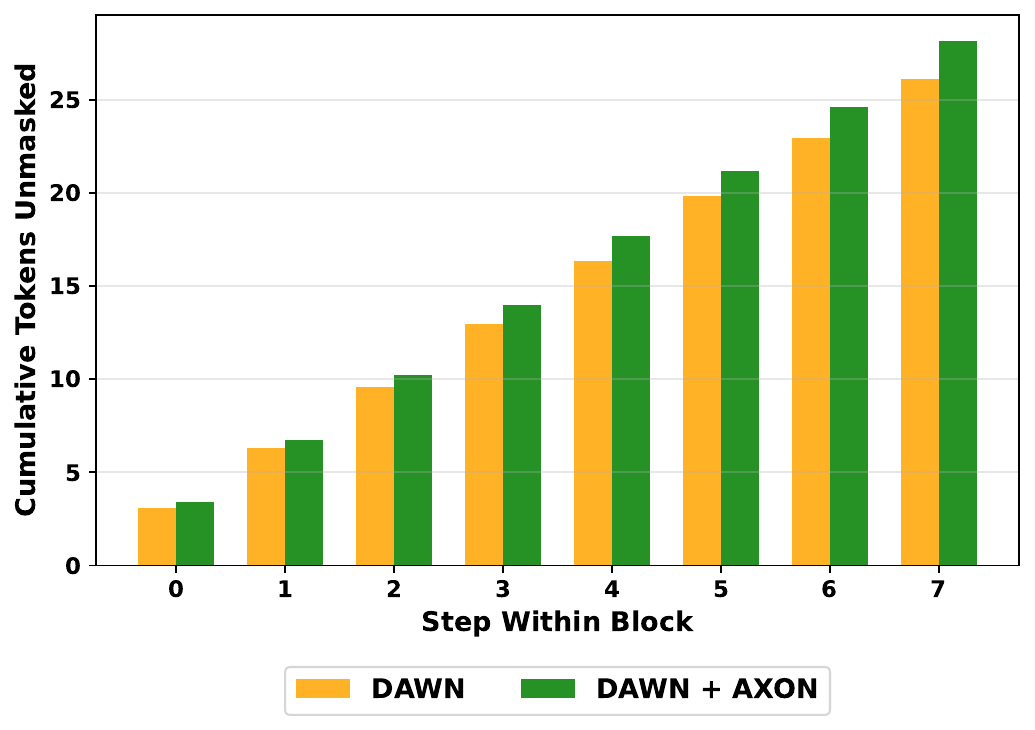}
    \caption{Cumulative tokens unmasked}
    \label{fig:unmask_cumul}
  \end{subfigure}
  \hfill 
  \begin{subfigure}{0.48\linewidth}
    \centering
    \includegraphics[width=\linewidth]{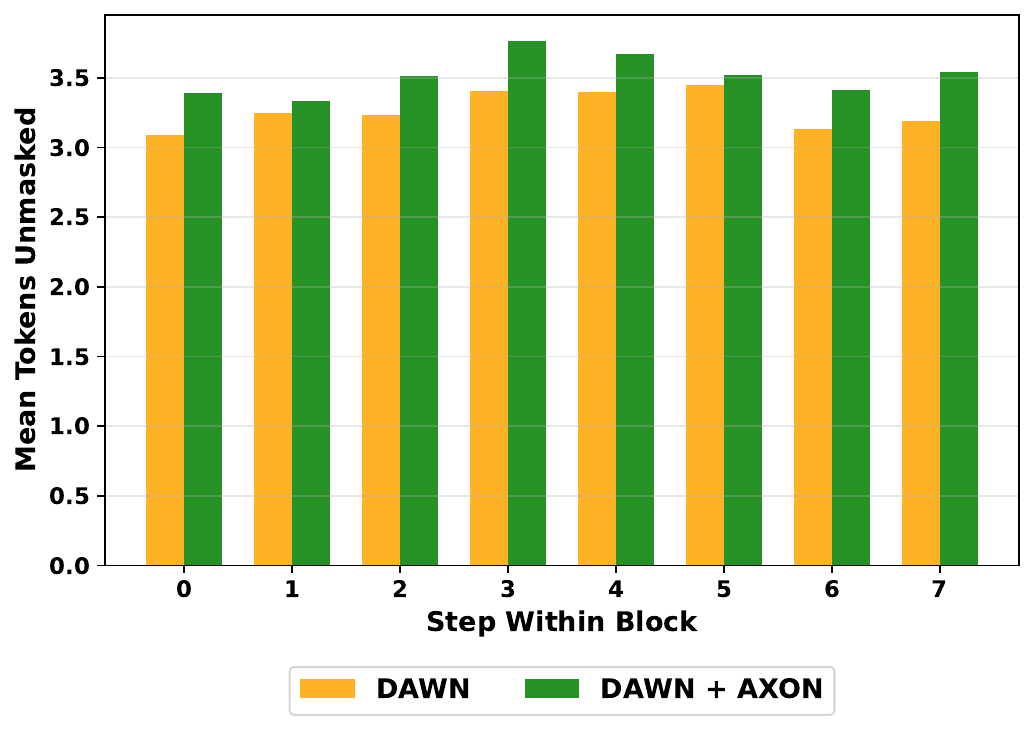}
    \caption{Mean tokens unmasked per individual step}
    \label{fig:unmask_perstep}
  \end{subfigure}

  \caption{\textbf{Token unmasking dynamics.} Analysis of generation mechanics within a single decoding block on Humaneval dataset, utilizing LlaDA1.5, and on top of DAWN, comparing (a) the cumulative trajectory of resolved tokens and (b) the mean unmasking rate per discrete step. \method{} consistently unmask a larger number of tokens per step.}
  \label{fig:unmasking_dynamics_combined}
\end{figure*}

To isolate \method{}'s efficiency gains (NFE), we analyze the unmasking profiles within a single decoding block in Figure~\ref{fig:unmasking_dynamics_combined}. Both DAWN and AXON maintain a remarkably flat per-step profile (Figure~\ref{fig:unmask_perstep}). However, \method{} consistently achieves a stable $\sim$9\% advantage at every step, resolving roughly $3.5$ tokens compared to DAWN's $3.2$. As shown in the cumulative trajectory (Figure~\ref{fig:unmask_cumul}), this minor dividend systematically compounds, allowing \method{} to resolve $28/32$ tokens by the end of the block versus DAWN's $26/32$. Aggregated over a full decoding horizon, this saves roughly 16 forward passes within the same token budget, directly explaining \method{}'s lower overall NFE. The consistent gap, caused by \method{} firing  , confirms that submodular coverage steadily guides the model toward high-yield positions.

\section{Submodularity of the Coverage Objective}
\label{app:coverage-submodular-proof}

\newcounter{coverageproposition}
\refstepcounter{coverageproposition}
\noindent\textbf{Proposition~\thecoverageproposition.}
\label{prop:coverage-submodular}
The facility-location coverage function in Eq.~\ref{eq:coverage-objective}
is monotone submodular.

\noindent\textbf{Proof.}
Fix a denoising step $t$, and let
$U$ be the residual positions that the anchor set should support and
$V$ be the candidate anchor positions. Assume all weights are
nonnegative, $w_{ij}\ge 0$, which holds for
$w_{ij}^{(t)}=A_{ij}^{(t)}(1-c_i^{(t)})c_j^{(t)}$ because attention
weights and confidence terms are nonnegative. For any selected anchor
set $S\subseteq V$, recall that
\[
    C(S)=\sum_{i\in U}\max_{j\in S} w_{ij},
\]
with the convention that $\max_{j\in\emptyset}w_{ij}=0$.

\paragraph{Monotonicity.}
Let $A\subseteq B\subseteq V$. For each residual position $i$,
\[
    \max_{j\in A}w_{ij}\le \max_{j\in B}w_{ij},
\]
because $B$ contains every candidate in $A$ and possibly more. Summing
over $i\in U$ gives $C(A)\le C(B)$. Therefore, $C$ is monotone.

\paragraph{Submodularity.}
We show diminishing returns. Let $A\subseteq B\subseteq V$ and let
$e\in V\setminus B$. For any set $S\subseteq V$, write
\[
    m_i(S)=\max_{j\in S}w_{ij}.
\]
The marginal gain from adding $e$ to $S$ is
\[
\begin{aligned}
    C(S\cup\{e\})-&C(S)
    \\
    &=\sum_{i\in U}
    \left(\max\{m_i(S),w_{ie}\}-m_i(S)\right) \\
    &=
    \sum_{i\in U}\max\{0,w_{ie}-m_i(S)\}.
\end{aligned}
\]
Since $A\subseteq B$, we have $m_i(A)\le m_i(B)$ for every $i\in U$.
Thus,
\[
    \max\{0,w_{ie}-m_i(A)\}
    \ge
    \max\{0,w_{ie}-m_i(B)\}
\]
for every $i$. Summing over all residual positions yields
\[
    C(A\cup\{e\})-C(A)
    \ge
    C(B\cup\{e\})-C(B).
\]
This is exactly the submodularity condition. Hence $C$ is monotone
submodular. \hfill$\square$

\section{Submodular-Objective Ablation}
\label{app:submod-ablation}
We ablate \method{}'s set-level anchor-selection objective along two axes.
\begin{enumerate}
\setlength{\itemsep}{0em}
\item \textbf{Coverage form.} Facility-location (FL) $C^{(t)}_{\mathrm{FL}}(S)=\sum_{i\in U}\max_{j\in S}w_{ij}^{(t)}$, in which each uncertain position draws its coverage from the best-matching anchors within the selected set $S$, versus Graph-Cut (GC) $C^{(t)}_{\mathrm{GC}}(S)=\sum_{i\in U}\sum_{j\in S}w_{ij}^{(t)}$, which instead sums each position's affinity over all selected anchors.
\item \textbf{Monotonicity.} The monotone objective $F_{\mathrm{mono}}(S)=C^{(t)}(S)$ versus the non-monotone $F_{\mathrm{nonmono}}(S)=C^{(t)}(S)-\lambda\,R(S)$, where $R(S)=\sum_{\{j,k\}\subseteq S} q_{jk}$ is a pairwise redundancy/conflict penalty over the selected set and its weight, $\lambda\ge 0$.
\end{enumerate}

\paragraph{Protocol.}
All ablation runs use the standard \method{} configuration. We sweep on the two datasets. \textsc{MBPP} on all three models, and HumanEval on the two LLaDA models, giving five (model, task) cells. For each cell, we run both objectives in \{$\mathrm{FL}$, $\mathrm{GC}$\}$\times$\{mono, nonmono\}, and for the non-monotone variants we sweep $\lambda\in\{0.2, 0.3, 0.4, 0.5\}$, giving $10$ configurations per cell.

\subsection*{Objective form: FL vs.\ GC (monotone)}
\label{app:fl-vs-gc}

We first compare the monotone facility-location (FL) objective against a graph-cut (GC). As we showed in Table~\ref{tab:fl-vs-gc}, FL consistently outperforms GC in computational efficiency while maintaining comparable accuracy. FL achieves a lower number of function evaluations (NFE) across all five evaluation cells, yielding a lower mean NFE overall (\textit{44.89} vs.\ \textit{45.95}). This confirms that explicitly enforcing structural diversity prevents the model from generating additive and redundant anchor coverage, thus streamlining the generation process without sacrificing task performance.

\subsection*{Non-monotone penalty: $\lambda$ sweep on FL}
\label{app:lambda-sweep}

We test whether explicitly penalizing redundant anchors improves selection. As shown in Table~\ref{tab:lambda-sweep}, we sweep the penalty weight $\lambda$ on the FL objective to observe if non-monotone regularization yields a better accuracy-NFE tradeoff compared to the purely monotone baseline ($\lambda=0$). The results demonstrate that introducing a non-monotone redundancy penalty provides no meaningful benefit to either task accuracy or generation efficiency. Across all five evaluation cells, shifting $\lambda$ from $0$ up to $0.5$ leaves the mean accuracy virtually unchanged (fluctuating negligibly between \textit{39.88}\% and \textit{39.96}\%). Similarly, the computational cost remains entirely stagnant, with the mean NFE locked tightly between \textit{44.86} and \textit{44.91}. This flat trend empirically justifies our choice of the purely monotone objective: because complex pairwise coordination penalties do not yield a superior improvement, the simpler, hyperparameter-free, monotone objective ($\lambda=0$) remains the optimal configuration.

\begin{table*}[t]
\centering

\footnotesize
\setlength{\tabcolsep}{6pt}
\begin{tabular}{ll rr rr rr rr rr}
\toprule
& & \multicolumn{2}{c}{$\lambda{=}0$ (mono)}
  & \multicolumn{2}{c}{$\lambda{=}0.2$}
  & \multicolumn{2}{c}{$\lambda{=}0.3$}
  & \multicolumn{2}{c}{$\lambda{=}0.4$}
  & \multicolumn{2}{c}{$\lambda{=}0.5$}\\
\cmidrule(lr){3-4} \cmidrule(lr){5-6} \cmidrule(lr){7-8} \cmidrule(lr){9-10} \cmidrule(l){11-12}
\textbf{Model} & \textbf{Task} & Acc & NFE & Acc & NFE & Acc & NFE & Acc & NFE & Acc & NFE \\
\midrule
Dream-7B  & MBPP      & 55.00 & 41.81 & 55.00 & 41.82 & 54.80 & 41.88 & 54.80 & 41.90 & 54.80 & 41.91 \\
LLaDA-1.5 & MBPP      & 35.20 & 28.30 & 35.40 & 28.19 & 35.40 & 28.19 & 35.40 & 28.18 & 35.20 & 28.19 \\
LLaDA-8B  & MBPP      & 30.40 & 42.40 & 30.40 & 42.40 & 30.40 & 42.41 & 30.40 & 42.42 & 30.40 & 42.60 \\
LLaDA-1.5 & HumanEval & 39.02 & 59.49 & 39.02 & 59.43 & 39.02 & 59.43 & 39.02 & 59.30 & 39.02 & 59.30 \\
LLaDA-8B  & HumanEval &  39.98 & 52.43 &  39.98 & 52.59 &  39.98 & 52.65 &  39.98 & 52.65 &  39.98 & 52.29 \\
\midrule
\multicolumn{2}{l}{Mean across cells}
  & \textit{39.92} & \textit{44.89}
  & \textit{39.96} & \textit{44.89} 
  & \textit{39.92} & \textit{44.91} 
  & \textit{39.92} & \textit{44.89} 
  & \textit{39.88} & \textit{44.86} \\
\bottomrule
\end{tabular}
\caption{\textbf{Ablating monotonicity.} Results comparing facility-location (FL) with monotone objective ($\lambda=0$) against the nonmonotone objective ($\lambda\in\{0.2,0.3,0.4,0.5\}$) across tasks and model configurations.}
\label{tab:lambda-sweep}
\end{table*}

\begin{figure}
    \centering
    \includegraphics[width=1.0\linewidth]{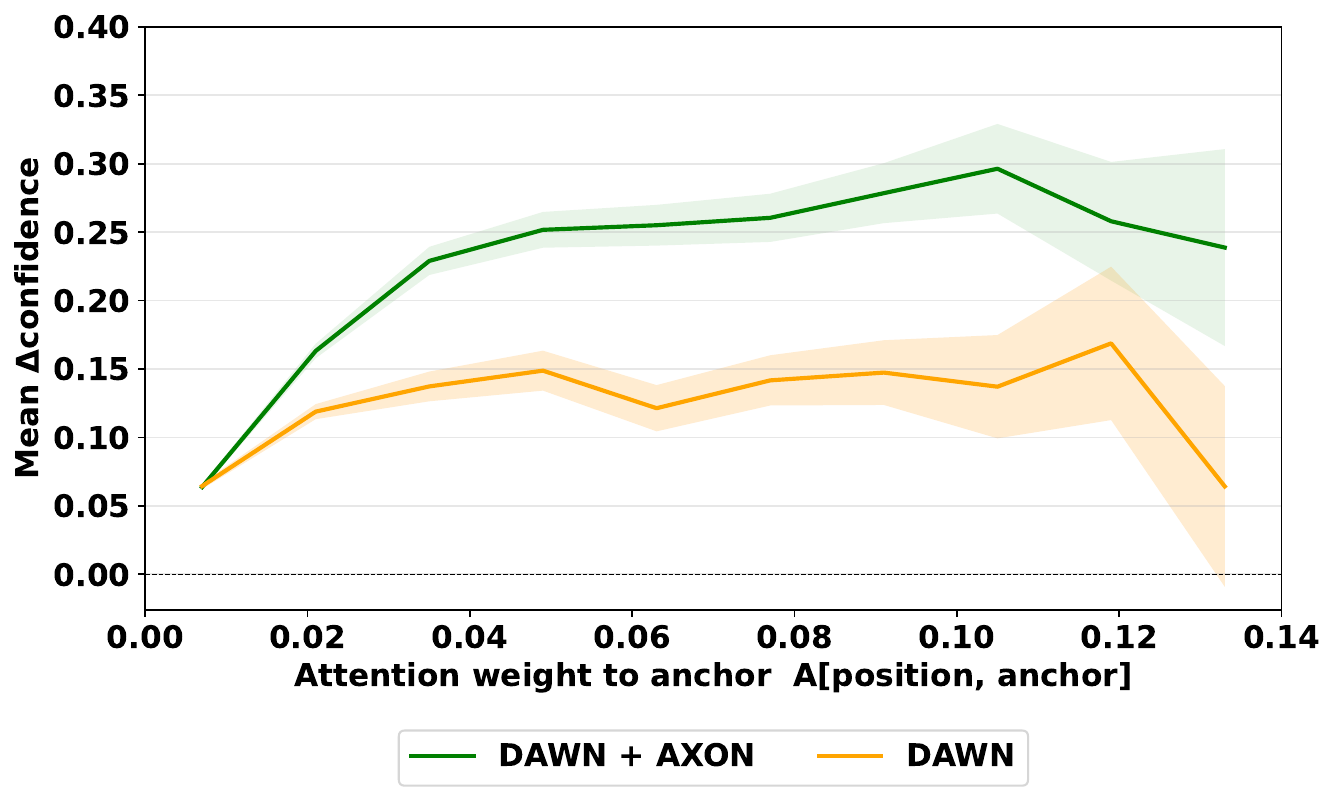}
    \caption{\textbf{Confidence propagation from \method{} anchors scales with attention connectivity.} 
For each decoding step where a submodular anchor token is committed, we track the per-position confidence gain $\Delta\text{confidence} = x^p_0[t+1] - x^p_0[t]$ for all remaining masked positions, binned by their attention weight to that anchor $A[\text{position}, \text{anchor}]$. Results are averaged over HumanEval dataset, and utilizing LLaDA-1.5.}
    \label{fig:confidence_attention2anchor}
\end{figure}

\subsection*{Confidence Propagation From \method{}} Figure~\ref{fig:confidence_attention2anchor} validates the core structural premise of \method{}'s facility-location objective: anchors should be chosen not merely for their own high confidence, but for their structural capacity to propagate information to the unmasked positions that attend to them.  At near-zero attention connectivity ($A \approx 0$), both methods perform identically, yielding a minimal confidence gain ($\sim 0.04$). 
However, as attention weight increases, \method{}'s confidence gain ($\Delta\text{confidence}$) rises steeply and plateaus at $\sim 0.25\text{--}0.30$. 
This represents a 2x improvement over DAWN's flat plateau ($\sim 0.12\text{--}0.15$), proving that \method{}'s submodular selection successfully routes information to highly-attending positions to accelerate model confidence.

\section{Gate Weight \texorpdfstring{$\alpha$}{alpha} Ablation}
\label{apendix:alpha-ablation}

The hybrid gate of \method{} combines two deficit signals through a single weight $\alpha$: the context deficit $d_{\text{ctx}}^{(t)}$ Eq.~\ref{eq:dcov} and the pace deficit $d_{\text{pace}}^{(t)}$ Eq.~\ref{eq:dpace}. It fires whenever the convex combination of their positive rectifications exceeds zero (see Eq.~\ref{eq:Gate}). The two endpoints recover the pure regimes: $\alpha\!=\!1$ gates solely on the context deficit (ensuring contextual support scales with residual uncertainty and commit volume), while $\alpha\!=\!0$ gates solely on the pace deficit (ensuring the base decoder maintains the required completion schedule). 

\begin{figure*}[t]
  \centering
  
  \begin{subfigure}{0.32\linewidth}
    \centering
    \includegraphics[width=\linewidth]{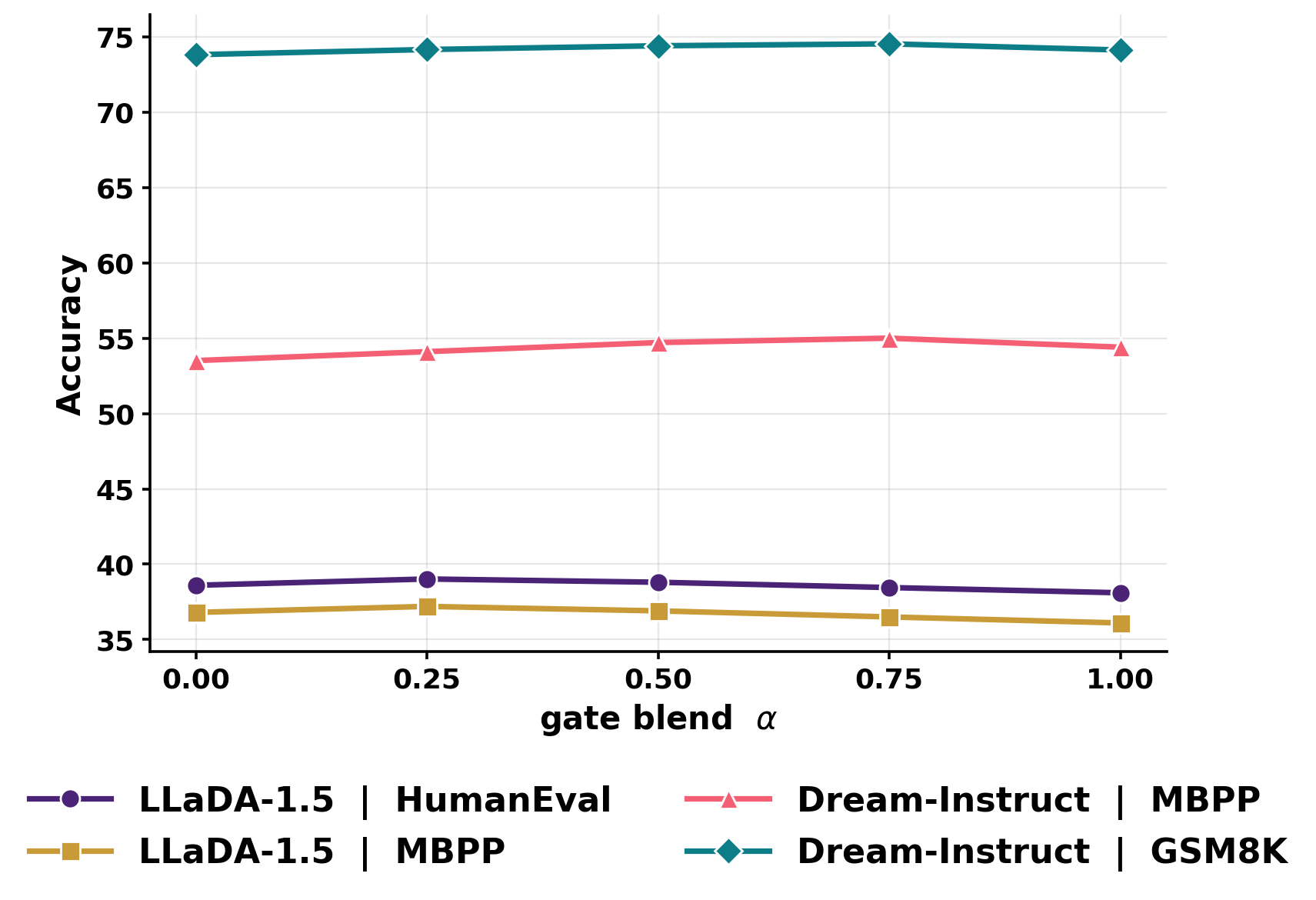}
    \caption{Accuracy vs.\ $\alpha$}
    \label{fig:cp_alpha_acc}
  \end{subfigure}
  \hfill 
  \begin{subfigure}{0.32\linewidth}
    \centering
    \includegraphics[width=\linewidth]{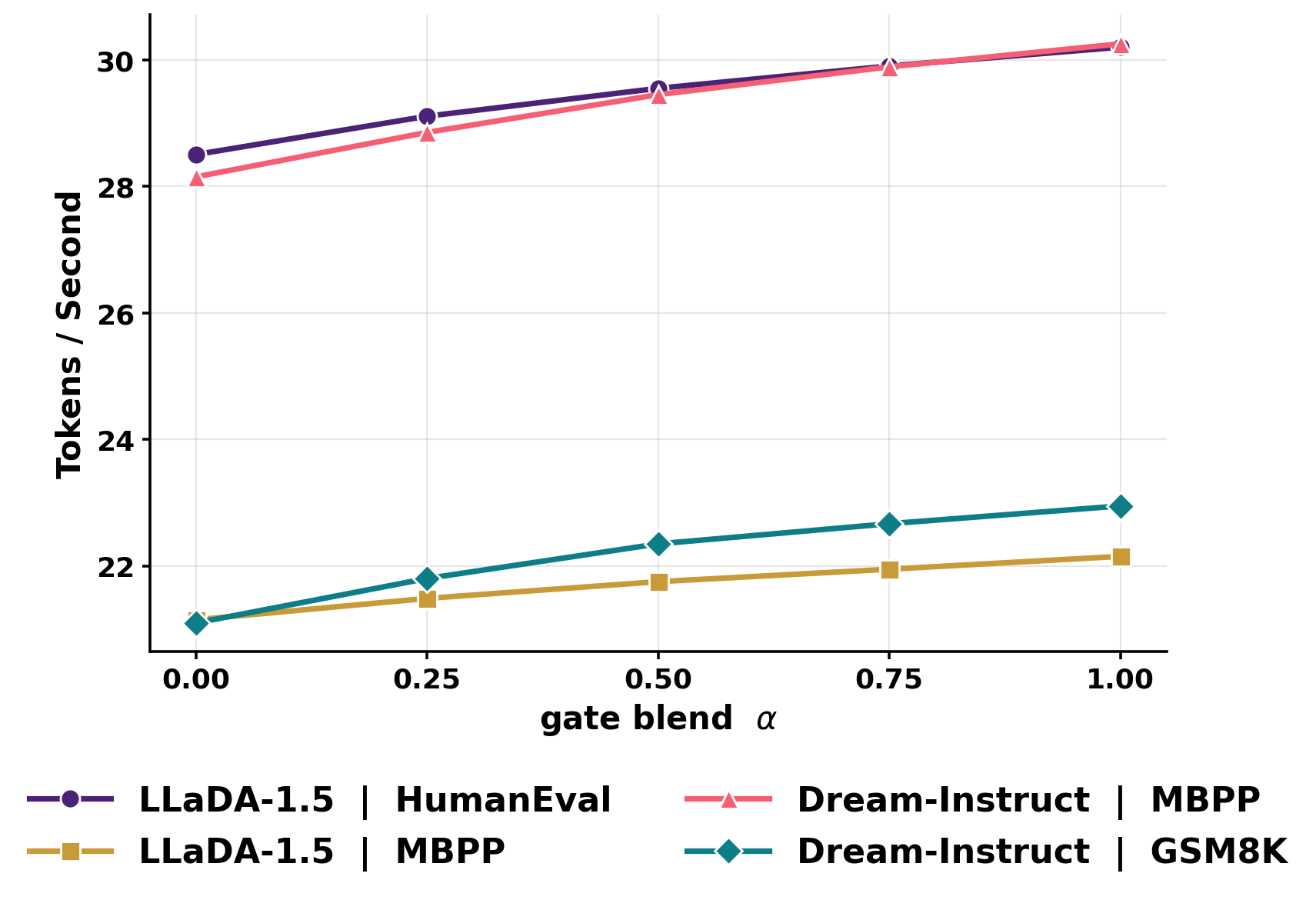}
    \caption{Throughput vs.\ $\alpha$}
    \label{fig:cp_alpha_tps}
  \end{subfigure}
  \hfill 
  \begin{subfigure}{0.32\linewidth}
    \centering
    \includegraphics[width=\linewidth]{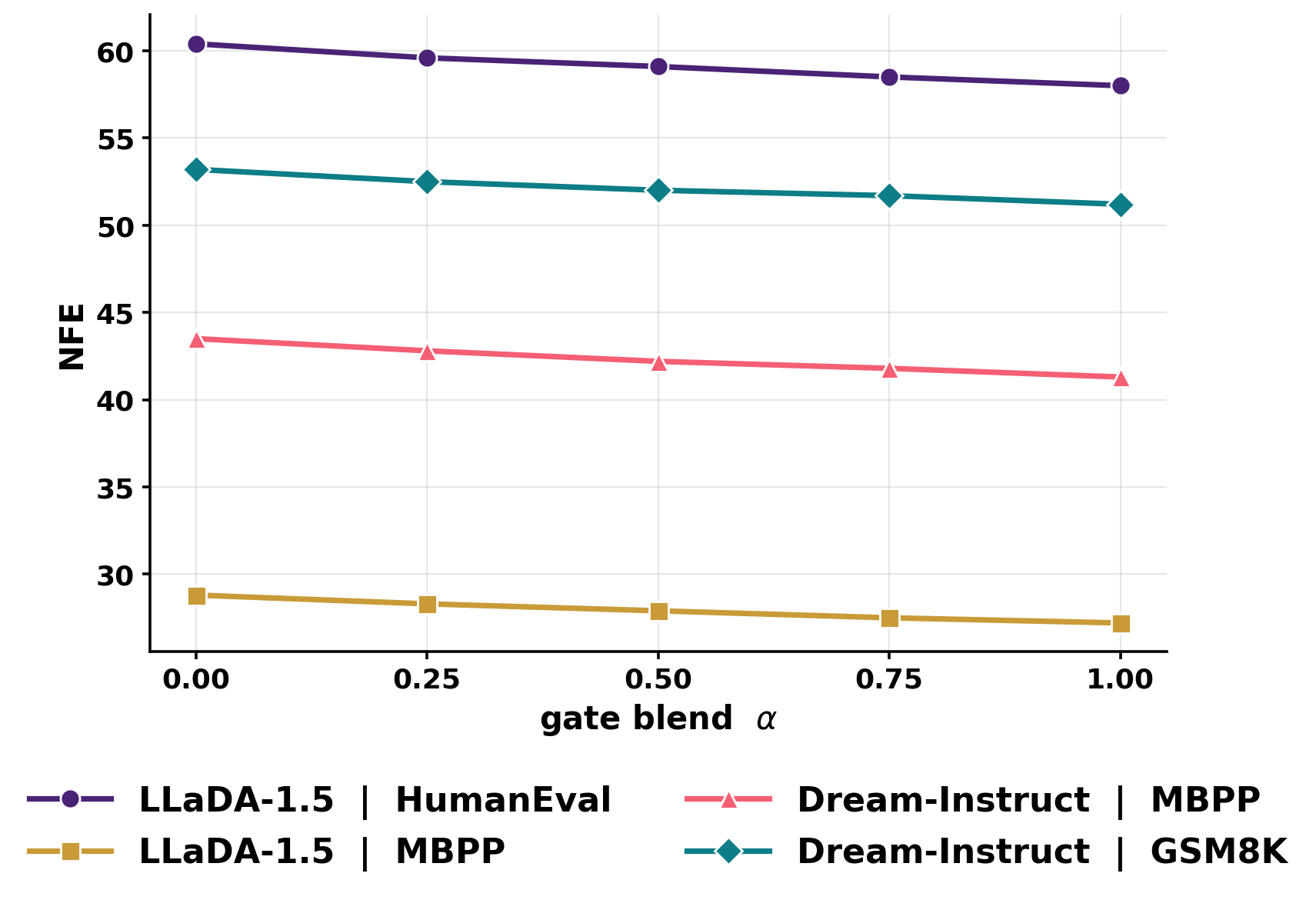}
    \caption{Average NFE vs.\ $\alpha$}
    \label{fig:cp_alpha_nfe}
  \end{subfigure}

  \caption{\textbf{Ablating gate weight $\alpha$.} Performance metrics across multiple models and tasks evaluated on top of DAWN under varying gate configurations, $\alpha \in \{0.0, 0.25, 0.5, 0.75, 1.0\}$. This illustrates the simultaneous impact of $\alpha$ on (a) downstream task accuracy, (b) token generation throughput (TPS), and (c) total computational cost measured in average NFE.}
  \label{fig:cp_alpha_combined}
\end{figure*}

To quantify how sensitive \method{} is to this choice, we sweep $\alpha \in \{0,\,0.25,\,0.5,\,0.75,\,1\}$ on top of the DAWN base decoder (Figure~\ref{fig:cp_alpha_combined}). We evaluate four model $\times$ task configurations: LLaDA-1.5 on HumanEval and MBPP, and Dream-Instruct on MBPP and GSM8K, reporting accuracy, TPS, and NFE as functions of $\alpha$ in Figures~\ref{fig:cp_alpha_acc}, \ref{fig:cp_alpha_tps}, and~\ref{fig:cp_alpha_nfe} respectively.

\paragraph{Accuracy.} As illustrated in Figure~\ref{fig:cp_alpha_acc}, the impact of the gate weight $\alpha$ on generation accuracy reveals a clear, model-specific optimum, demonstrating that performance is non-monotonic. Pushing the gate to either extreme relying solely on the pace schedule ($\alpha = 0$) or exclusively on contextual support ($\alpha = 1.0$) is often sub-optimal, validating the need to interpolate between both deficit checks. We configure \method{} with $\alpha=0.2$ and $\alpha=0.9$ for the LLaDA and Dream-Instruct backbones respectively across all main-text evaluations, tuning the framework to the specific failure modes of each base model.

\paragraph{TPS Throughput and NFE.} Generation throughput and computational overhead scale dynamically with $\alpha$ (see Figures~\ref{fig:cp_alpha_tps} and \ref{fig:cp_alpha_nfe}). By shifting the gate's sensitivity, the framework becomes more or less tolerant of rapid, under-supported commits, thereby firing the submodular selection at different frequencies. Tuning this parameter safely accelerates generation speed and streamlines compute cost, offering a predictable trade-off trajectory against downstream quality depending on the strictness of the deployment budget.

\section{Relationship between confidence gain and attention}
\label{app:confgain_attention_theory}

\paragraph{Overview and bound.}In this section we analyze how revealing a chosen anchor affects the confidence of the model at a masked position. Explicitly, we will prove an upper bound on the increase in confidence at a masked position after the anchor is revealed, in terms of the attention map $A$. Let the masked position be  $i$, and the chosen anchor be at position $s$ and let $y_s \in \mathcal{V} \setminus \{\mask\}$ be the revealed (anchor) token. For simplicity, we assume that the prompt $X$ is empty, i.e. the input to the dLLM consists of only the response vector $y^{(t)} = (y_1^{(t)}, \ldots, y_L^{(t)})$. Note that $y^{(t)}_s = \mask$ prior to revealing the anchor. Denote the response vector after the reveal as $\tilde{y}^{(t)}$, where
\begin{align*}
    \tilde{y}^{(t)}_j = \begin{cases}
        y_s \quad \text{if } j=s, \\
        y^{(t)}_j  \quad \text{otherwise .} \\
    \end{cases}
\end{align*}
Let $\mathcal{V}'= \mathcal{V} \setminus \{\mask\}$. Thus, the increase in confidence (\textit{confidence gain}) for position $i$ after revealing position $s$ can be defined as,
\begin{align*}
    \delta_i = \max_{v \in \mathcal{V}'} p_{\theta}(y_i = v | \tilde{y}^{(t)}) -   \max_{v \in \mathcal{V}'} p_{\theta}(y_i = v | y^{(t)}).
\end{align*}
While it is intuitive that the confidence of position $i$ should be more sensitive to the reveal of token $s$ when the corresponding attention entry $A_{is}$ is large, it is challenging to make this intuition rigorous due to the complexity of the transformer model $p_{\theta}$. To make progress in this direction, we adopt certain simplifying assumptions following prior work \cite{zhou2026attention}, namely, 
\begin{enumerate}
    \item Following \cite{zhou2026attention}, we assume that the model $p_{\theta}$ is parametrized by a single layer transformer.
    \item We assume that the attention map $A$ does not change after revealing position $s$ (similar to \cite{zhou2026attention} Assumption 3.1). 
\end{enumerate}
Under these assumptions, we can prove that, 
\begin{align}
    \delta_i \leq \beta A_{is}, \label{eqn:confidence_upper_bound}
\end{align}
for a positive constant $\beta > 0$.

\paragraph{Discussion of bound.} The upper bound in \eqref{eqn:confidence_upper_bound} motivates our use of attention as a measure for anchor usefulness, and the inclusion of $A_{ij}$ in the score $w_{ij}$. Indeed, \eqref{eqn:confidence_upper_bound} implies that the improvements for masked tokens (with respect to confidence) is limited by their dependency on the anchor position $s$ (as measured by $A_{is})$. This is corroborated by Figure \ref{fig:confidence_attention2anchor}, where low attention values are correlated with low confidence gains. Thus, naively choosing an anchor $s$ may have only a minimal effect on the confidence of remaining tokens if the column $A_{\cdot, s}$ has small entries. This potentially explains the superior performance of \method{} with respect to confidence gain (see Figure \ref{fig:confidence_attention2anchor}), since 
\method{}'s submodular objective explicitly addresses this factor. 

However, our analysis is limited by our assumptions. In particular, assuming a static attention map fails to account for the case where revealing an anchor $s$ strongly affects a masked position $i$, that was earlier coupled with the anchor only through a positional dependency (e.g. adjacent tokens). For example, if the anchor is revealed to be a word that primarily occurs in a multi-token entity, e.g. ``York'', the attention map can potentially change a lot for certain positions (e.g. the preceding position). Extending this analysis to consider such effects is left for future work. We now explicitly discuss the assumed architecture of $p_{\theta}$, followed by a proof of \eqref{eqn:confidence_upper_bound}.

\paragraph{Architecture assumption.} Explicitly, let $\text{Embed}: \mathcal{V} \to \mathbb{R}^d$ be a d-dimensional embedding of the token space, and $\text{Positional} : \{1, \ldots, L\} \to \mathbb{R}^d$ be a positional encoding function. As before, let $y^{(t)} = (y_1, \ldots, y_L)$ be the partially masked response at diffusion step $t$. Let $p_{\theta}(y_i| y^{(t)}) \in \mathbb{R}^{|\mathcal{V}|}$ denote the full categorical distribution predicted by the model. 

For our assumed architecture, the predicted distribution is computed in the following steps. First, the embeddings for each position $i$ are computed as follows, 
\begin{align*}
    h_i = \text{emb}(y_i, i) = \text{Embed}(y_i) + \text{Positional}(i).
\end{align*}
Given the query, key and value projection matrices $W_{Q}, W_{K}, W_V$, the attention maps $A_{ij}$ and output representations for the attention block are computed as follows, 
\begin{align*}
    &q_i = W_Q h_i, \quad  k_i = W_K h_i, \quad v_i= W_V h_i, \\
    &A_{ij} = \frac{\text{exp}(q_i^\top k_j/\sqrt{d})}{\sum_j \text{exp}(q_i^\top k_j/\sqrt{d})},\\
    & z_i = \sum_{j} A_{ij} v_j.
\end{align*}
Finally, a fully connected layer (with parameters $(W,b)$), followed by a softmax activation, is applied to each position independently  to obtain the predicted distributions as follows :
\begin{align*}
    p_{\theta}(y_i| y^{(t)})  = \text{Softmax}(W z_i + b). 
\end{align*}
Define the notation,
\begin{align*}
    g(z) =  \text{Softmax}(W z + b).
\end{align*}
Since $g$ is differentiable everywhere, therefore $g$ is also locally Lipschitz continuous. Note that since $\mathcal{V}$ and $\{1, \ldots, L\}$ are finite sets, given that $\text{emb}(v,i)$ is always well-defined, there exists some $B_e >0$ such that $\|h_i\|_2 \leq B_e$ for every position $i$. Next, since $W_V$ is a (bounded) linear mapping and $A_{ij} \in [0,1]$, there also exists some $B > 0$ such that $\|z_i\|_2 \leq B$ for every position $i$.  Since $g$ is locally Lipschitz, it is globally Lipschitz on the bounded ball $\|z\|_2 \leq B$. Let this global Lipschitz constant be $L$, therefore, 
\begin{align}
       \|g(z_1) - g(z_2)\|_1 \leq L \|z_1 -z_2\|_2.  \label{eqn:lipschitz_assumption}
\end{align}
Note that we use the absolute value norm on the left hand side. Since all norms are topologically equivalent in Euclidean spaces, \eqref{eqn:lipschitz_assumption} is equivalent to expressing it purely in terms of the Euclidean norm on both sides. This Lipschitz property is the main tool we will use to prove the upper bound \eqref{eqn:confidence_upper_bound} in the next part. 

\paragraph{Derivation of upper bound. } 
Clearly, 
\begin{align*}
    &\phantom{\leq} \delta_i \\
    &\leq | \delta_i| \\
            &=  | \max_{v \in \mathcal{V}'} p_{\theta}(y_i = v | \tilde{y}^{(t)}) -   \max_{v \in \mathcal{V}'} p_{\theta}(y_i = v | y^{(t)}) |\\
            &\leq \max_{v \in \mathcal{V}'} | p_{\theta}(y_i = v | \tilde{y}^{(t)}) -  p_{\theta}(y_i = v | y^{(t)}) |,
\end{align*}
where the last bound follows from the identity $|\max_{x} f(x)  - \max_x g(x)| \leq \max_x |f(x) - g(x)|$. Next, clearly the maximum entry of a positive valued vector is smaller than the sum of its entries, therefore, 
\begin{align*}
    &\max_{v \in \mathcal{V}'} | p_{\theta}(y_i = v |  \tilde{y}^{(t)}) -  p_{\theta}(y_i = v | y^{(t)}) |\\
    \leq &\|p_{\theta}(y_i |  \tilde{y}^{(t)}) -   p_{\theta}(y_i = v | y^{(t)})\|_1 \\
    =&\|g(\tilde{z}_i) - g(z_i)\|_1, 
\end{align*}
where $\tilde{z}_i$ is the output representation at position $i$ for the input $\tilde{y}^{(t)}$ (i.e. after unmasking the anchor).  We can thus apply the Lipschitz bound \eqref{eqn:lipschitz_assumption} to obtain, 
\begin{align*}
    \delta_i &\leq L \| \tilde{z}_i - z_i \|_2 \\
            &= L \left\| \sum_j A_{ij} (\tilde{v}_j -  v_j) \right\|_2, 
\end{align*}
where $\tilde{v}_j$ is the value computed at position $j$ for input $\tilde{y}^{(t)}$. Note that we have used the assumption here that the attention map remains the same for inputs $\tilde{y}^{(t)}$ and $y^{(t)}$. Since only position $s$ is different, therefore, $\tilde{v}_j = v_j$ iff $j \neq s$. Therefore, 
\begin{align*}
    \delta_i \leq L  A_{is} \| \tilde{v}_s -  v_s \|_2
\end{align*}
Let 
\begin{align*}
    C = \max_{\stackrel{x, y \in \mathcal{V}}{i, j \in \{1,\ldots, L\}}} \| W_V \text{emb}(x, i) - W_V \text{emb}(y, j)\|.    
\end{align*}
Clearly $C > 0$, and since $\mathcal{V}$ and $\{1,\ldots, L\}$ are finite sets, therefore $C < \infty$. Therefore, 
\begin{align*}
     &\| \tilde{v}_s -  v_s \|_2 \\
     = &\|W_V \text{emb}(y_s, s) -  W_V \text{emb}(\mask, s) \|_2 \leq C.
\end{align*}
Thus, we obtain
\begin{align*}
     \delta_i \leq L C  A_{is},
\end{align*}
which proves \eqref{eqn:confidence_upper_bound}. 

\begin{figure*}[t]
\centering
\begin{minipage}{0.49\linewidth}
  \centering
  \includegraphics[width=\linewidth]{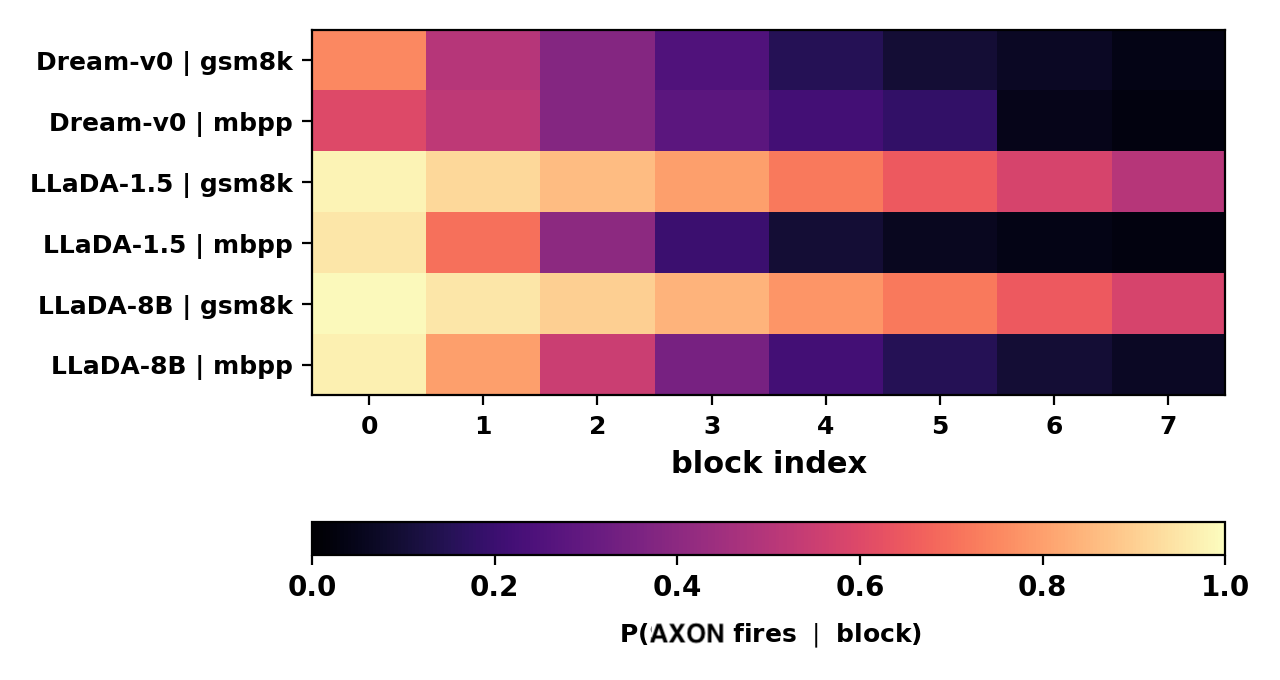}\\[2pt]
  \footnotesize\textbf{(a)} \textbf{\sage{} activation per block}
\end{minipage}
\hfill
\begin{minipage}{0.49\linewidth}
  \centering
  \includegraphics[width=\linewidth]{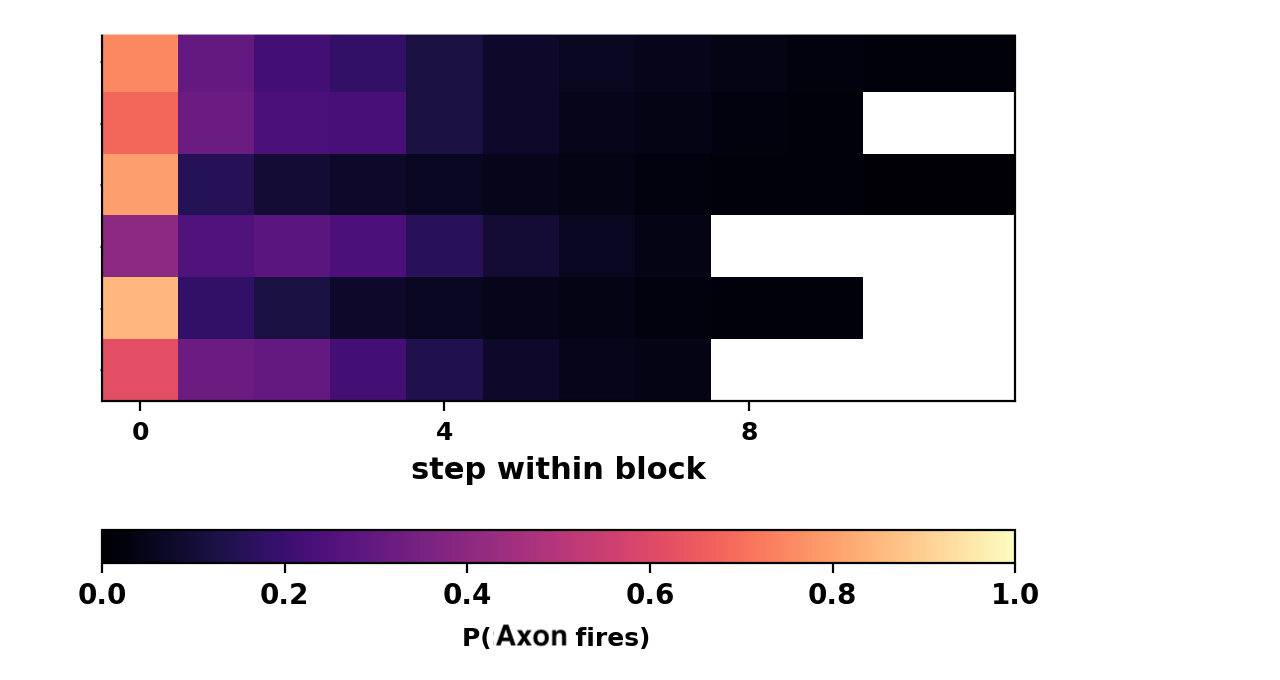}\\[2pt]
  \footnotesize\textbf{(b)} \textbf{\sage{} average activation per step}
\end{minipage}

\caption{\textbf{AXON gate firing patterns across generations.} 
Complementary heatmaps show the fraction of evaluation steps where the gate activates ($\mathrm{G}^{(t)}=1$) across Dream-Instruct and LLaDA backbones on MBPP and GSM8K. 
\textbf{(a) Within block activation.} \sage{} intervention heavily concentrate in the early blocks of a 256-token sequence, where the base proposer requires the most structural support. 
\textbf{(b) Within step activation.} Within any given block, \sage{} firing peaks during the first few denoising steps (at the block opening) and sharply declines once the base proposer catches up.}
\label{fig:firing-distribution}
\end{figure*}

\section{Effects Of \method{} On The Decoders}\label{sec:app_affect_on_decoders} 
The interaction between anchor selection and accelerated decoding strategies is heavily dependent on the structural representation mechanism of the method. 
For the simple thresholding-based methods, which make decisions to immediately commit a token only based on a flat probability threshold, \method{} acts as a co-dependency filter; by evaluating global information entropy, it penalizes mutual context redundancy, eliminating simultaneous-masking conflicts and yielding a steady recovery in both generation quality and Throughput.
For heuristically-driven spatial methods (LocalLeap), which expand unmasking zones natively by lowering confidence thresholds uniformly within a fixed, linear geometric radius around predicted anchors, \method{} enforces a diversity constraint that eliminates anchor clumping, preventing localized text degradation and primarily optimizing Accuracy. 
For graph-driven structural methods (DAWN), which is based on model token interdependencies by extracting raw attention maps and utilizing a Maximum Independent Set solver to parallelize non-conflicting tokens, \method{} provides global informational pillars that clarify noisy attention maps, allowing the dependency solver to uncover massive parallel sets and maximize throughput (TPS).

\section{Analysis of Gate Firing Patterns}
\label{app:gate_distribution}

Figure~\ref{fig:firing-distribution} visualizes the probability of the AXON gate firing ($\mathrm{G}^{(t)}=1$).

 \paragraph{Across blocks (Macro).} 
Figure~\ref{fig:firing-distribution}a shows that activations are heavily front-loaded. Early in the generation, causal context is sparse and residual uncertainty is high, requiring AXON to intervene frequently. As the sequence progresses and reliable context accumulates, the gate naturally backs off.

\paragraph{Within blocks (Micro).} 
Figure~\ref{fig:firing-distribution}b reveals that interventions concentrate almost entirely at Step 0. When a new block opens, structural support is minimal. AXON immediately injects high-leverage anchors to clear this bottleneck, allowing the base decoder to rapidly resolve the rest of the block uninterrupted. The truncated right edge of the heatmap visually confirms that this early intervention reduces the total steps required to finish the block.

\begin{figure}[t]
    \centering
    \includegraphics[width=1.0\linewidth]{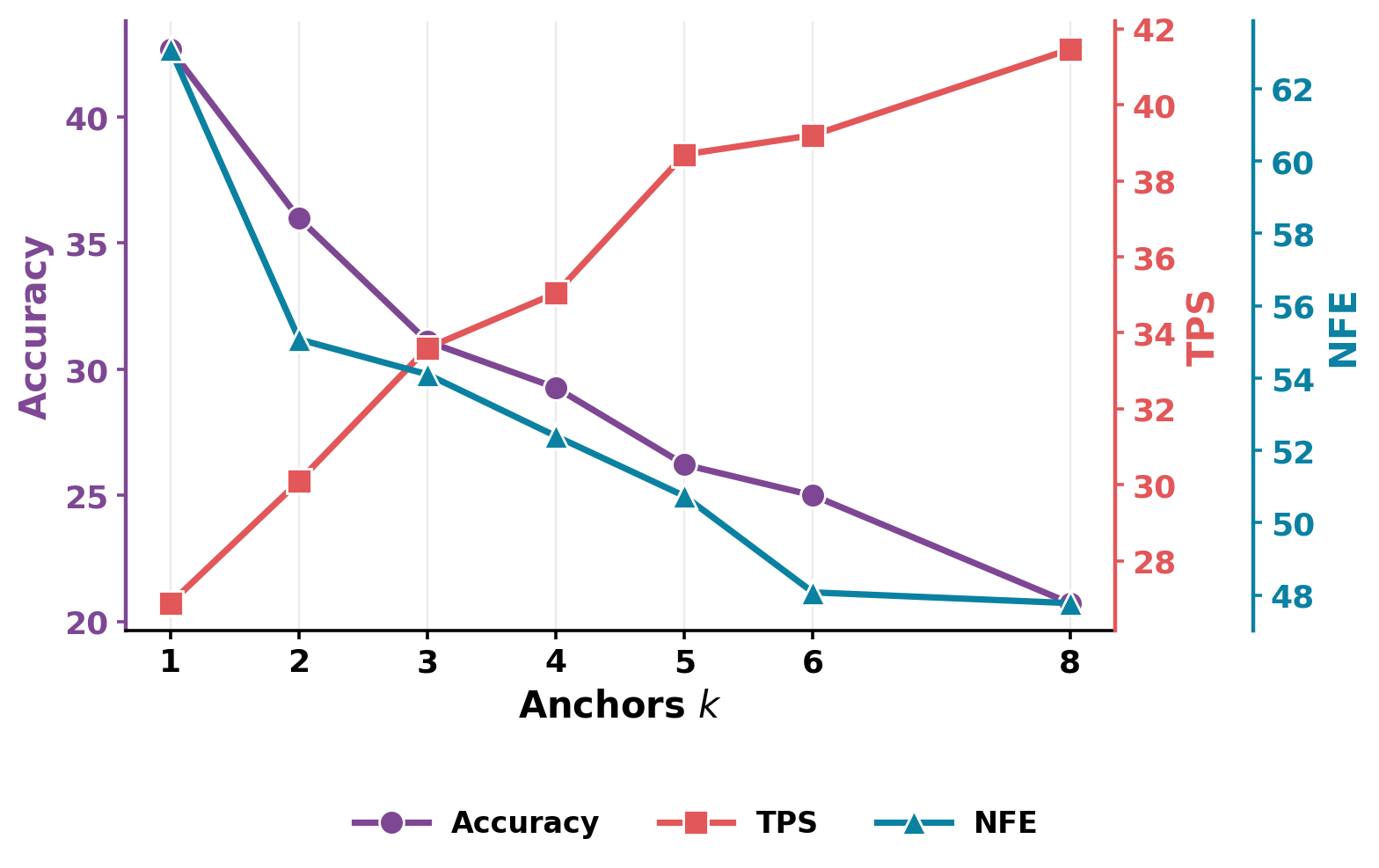}
    \caption{\textbf{Impact of anchor budget on decoding dynamics on Llada-1.5 and HumanEval.} Accuracy, throughput (TPS), and NFE across varying anchor set sizes.}
    \label{fig:anchor_ablation}
\end{figure}

\section{Effect of the Number of Anchors}
\label{app:anchor_ablation}




\method{} uses a small anchor budget because the goal is not to maximize the number of additional committed tokens, but to reveal a small number of tokens that provide high contextual support to uncertain residual positions. The default variant uses a fixed budget of $k=1$ anchor whenever the gate fires. This setting keeps the intervention cheap and conservative, while the facility-location objective discourages redundant reveals by giving each uncertain position credit only for its best selected anchor.

\begin{figure*}[t]
  \centering
  
  \begin{subfigure}{0.32\linewidth}
    \centering
    \includegraphics[width=\linewidth]{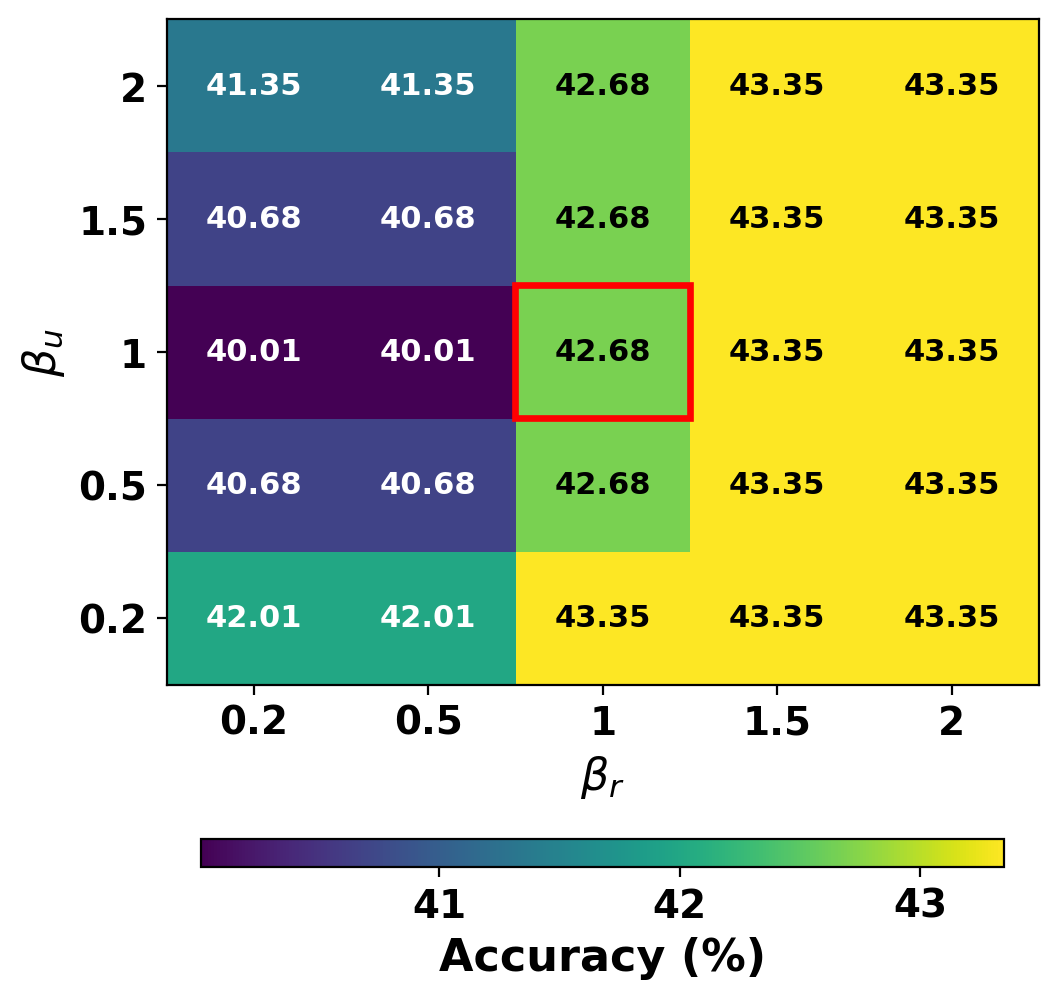}
    \caption{Accuracy vs. $\beta_u$ and $\beta_r$}
    \label{fig:cp_beta_acc}
  \end{subfigure}
  \hfill 
  \begin{subfigure}{0.32\linewidth}
    \centering
    \includegraphics[width=\linewidth]{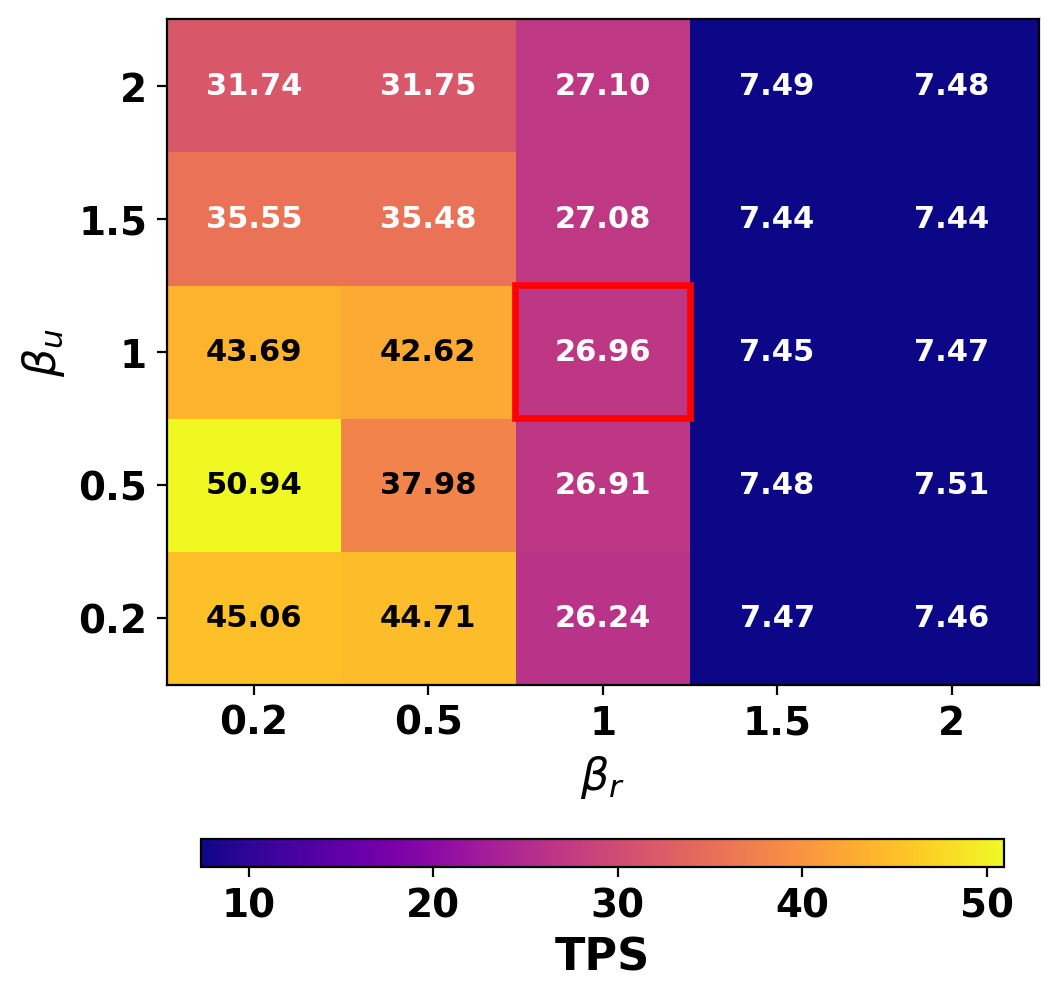}
    \caption{Throughput vs. $\beta_u$ and $\beta_r$}
    \label{fig:cp_beta_tps}
  \end{subfigure}
  \hfill 
  \begin{subfigure}{0.32\linewidth}
    \centering
    \includegraphics[width=\linewidth]{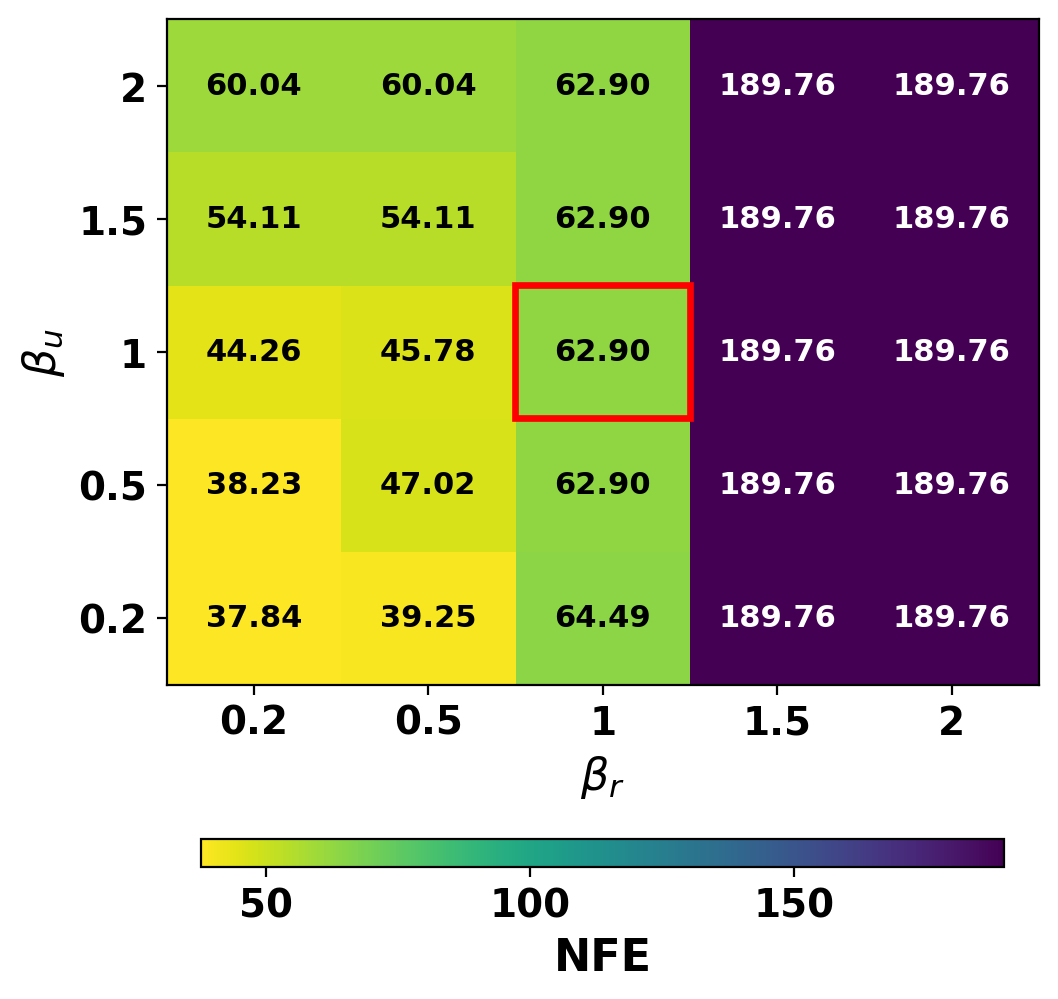}
    \caption{Average NFE vs. $\beta_u$ and $\beta_r$}
    \label{fig:cp_beta_nfe}
  \end{subfigure}

  \caption{\textbf{Ablating scaling coefficients $\beta_u$ and $\beta_r$.} Performance metrics utilizing LlaDa1.5 backbone, on HumanEval task, on top of DAWN decoder under varying scaling coefficients, $\{\beta_u,\beta_r\}\in\{0.2, 0.5, 1.0, 1.5, 2.0\} \times \{0.2, 0.5, 1.0,1.5,2.0\}$. This illustrates the simultaneous impact of $\beta_u$ and $\beta_r$ on (a) downstream task accuracy, (b) token generation throughput (TPS), and (c) total computational cost measured in average NFE.}
  \label{fig:cp_beta_combined}
\end{figure*}

We also evaluate \method{}\textsubscript{CVR}, an adaptive coverage-matching variant. Let
\[
\tau^{(t)} = C^{(t)}\!\left(S^{(t)}_{\mathrm{base}}\right)
\]
denote the contextual coverage induced by the base decoder's proposal. Starting from $S_0=\varnothing$, \method{}\textsubscript{CVR} greedily adds the candidate anchor with largest marginal gain under $C^{(t)}$ until
\[
C^{(t)}(S_m)\ge \tau^{(t)}.
\]
Equivalently, its final cardinality is
\[
|S^{\star(t)}_{\mathrm{CVR}}|
=
\min\left\{m\ge 1:\ C^{(t)}(S_m)\ge \tau^{(t)}\right\}.
\]

This gives the adaptive variant a per-step scale for how much contextual support is needed, while still selecting anchors according to the same non-redundant coverage objective. For any realized cardinality $m$, the greedy prefix retains the standard $(1-1/e)$ approximation guarantee for monotone submodular maximization under a cardinality budget of $m$.

In Figure~\ref{fig:anchor_ablation}, we ablate the fixed per-fire anchor budget and report its effect on NFE, tokens per second, and accuracy.

Larger budgets commit more tokens per intervention, shrinking the residual mask $M^{(t)}$ and lowering NFE monotonically. Throughput tracks NFE initially but plateaus once the greedy submodular search overhead begins to dominate the savings. Accuracy follows an inverted U-shape: a small set of high-confidence anchors absorbs the attention of uncertain residuals and resolves stalled positions, while larger budgets force selection from the low-confidence tail of the candidate pool $V^{(t)}$, injecting noise that propagates through subsequent denoising steps. The settings deployed in Section~\ref{sec:experiments} sit at the Pareto frontier of this trade-off.

\begin{table*}[t]
\centering
\footnotesize
\setlength{\tabcolsep}{5pt}
\begin{tabular}{ll rr r cc c}
\toprule
\textbf{Backbone} & \textbf{Decoder} & \textbf{plain} & \textbf{+\method{}} & $\Delta$\textbf{acc} & \textbf{NFE (plain$\to$+\method{})} & \textbf{TPS (plain$\to$+\method{})} & \textbf{Speedup} \\
\midrule
Dream-7B  & Confidence & 45.04 & 52.26 & +7.22 & 256.0 $\to$ 30.9 & 1.67 $\to$ 8.58  & 5.13$\times$ \\
          & DAWN       & 50.43 & 52.39 & +1.96 & 26.3 $\to$ 25.8  & 9.12 $\to$ 10.53 & 1.16$\times$ \\
          & LocalLeap  & 50.80 & 52.26 & +1.47 & 26.4 $\to$ 25.9  & 8.80 $\to$ 10.12 & 1.15$\times$ \\
\midrule
LLaDA-1.5 & Confidence & 50.92 & 51.16 & +0.24 & 256.0 $\to$ 36.9 & 1.37 $\to$ 9.36  & 6.81$\times$ \\
          & DAWN       & 49.82 & 50.31 & +0.49 & 33.9 $\to$ 27.4  & 10.36 $\to$ 12.79 & 1.23$\times$ \\
          & LocalLeap  & 50.80 & 50.92 & +0.12 & 37.0 $\to$ 31.5  & 9.27 $\to$ 10.66 & 1.15$\times$ \\
\bottomrule
\end{tabular}
\caption{\textbf{TruthfulQA.} Open-ended, non-coding, non-reasoning evaluation on Dream-7B and LLaDA-1.5.}
\label{tab:truthfulqa}
\end{table*}

\begin{table*}[t]
\centering
\footnotesize
\setlength{\tabcolsep}{4pt}
\begin{tabular}{l rrr rrr}
\toprule
& \multicolumn{3}{c}{\textbf{HumanEval}} & \multicolumn{3}{c}{\textbf{TruthfulQA}} \\
\cmidrule(lr){2-4} \cmidrule(lr){5-7}
\textbf{Decoder} & Acc. & TPS & NFE & Acc. & TPS & NFE \\
\midrule
baseline & 74.39 & 7.1 & 160.5 & 42.96 & 9.3 & 106.7 \\
\hdashline
Confidence & 76.22 & 7.7 & 164.7 & 43.33 & 8.5 & 109.6 \\
\quad +\method{} & 79.27 & 8.7 & 165.8 & 44.43 & 9.6 & 111.4 \\
\quad +\method{}\textsubscript{CVR} & 79.27 & 8.7 & 165.8 & 44.43 & 9.7 & 111.4 \\
\hdashline
LocalLeap & 73.78 & 8.4 & 150.6 & 43.45 & 8.9 & 104.9 \\
\quad +\method{} & 77.44 & 9.4 & 152.0 & 44.55 & 10.0 & 107.2 \\
\quad +\method{}\textsubscript{CVR} & 77.43 & 9.4 & 152.0 & 44.55 & 10.1 & 107.2 \\
\hdashline
DAWN & 74.39 & 9.1 & 148.5 & 43.70 & 10.7 & 105.3 \\
\quad +\method{} & 74.39 & 10.0 & 147.2 & 45.57 & 11.0 & 104.9 \\
\quad +\method{}\textsubscript{CVR} & 74.39 & 10.0 & 147.2 & 45.57 & 11.0 & 104.9 \\
\bottomrule
\end{tabular}
\caption{\textbf{SDAR-8B backbone results.} An additional dLLM backbone on HumanEval and TruthfulQA.}
\label{tab:sdar8b}
\end{table*}
\section{Context Deficit Normalization \texorpdfstring{$\beta$}{beta}}
\label{app:beta_ablation}

The context deficit $d_{\text{ctx}}^{(t)}$ (Eq.~\ref{eq:dcov}) evaluates the contextual support $g^{(t)}$ against two distinct signals: the decoding pace $r^{(t)}$ and the residual uncertainty $u_M^{(t)}$. While all three quantities naturally fall within the $[0, 1]$ interval, they originate from different distributions and lack a universally scaled theoretical equivalence. As discussed in Appendix~\ref{app:confgain_attention_theory}, this correlates between attention mass and uncertainty reduction. However, to account for potential distribution shifts across different model architectures, one could introduce scaling coefficients $\beta_r, \beta_u > 0$ to explicitly calibrate these thresholds:
\begin{equation}
d_{\text{ctx}}^{(t)}  = \max\!\big(\beta_r r^{(t)},\, \beta_u u_M^{(t)}\big) - g^{(t)}.
\end{equation}

Setting $\beta_r = \beta_u = 1$ establishes a parameter-free uniform-prior baseline which we use in our experiments in Table~\ref{tab:main}. To evaluate the sensitivity of \method{} to these normalization constants, we perform a 2D grid search over $\beta_r, \beta_u \in \{0.2, 0.5, 1.0, 1.5, 2.0\}$ (Figure~\ref{fig:cp_beta_combined}), reporting Accuracy, TPS, and NFE.
\paragraph{Accuracy.} 
In Figure~\ref{fig:cp_beta_acc}, the grid search reveals a clear trade-off spectrum depending on the strictness of the gate. At lower threshold values ($\beta \le 0.5$), the gate is overly permissive; it frequently fails to intervene when the base decoder lacks sufficient structural context, causing generation accuracy to drop. As the $\beta$ coefficients increase, the gate becomes stricter, forcing the model to rely more heavily on the injected anchors which pushes accuracy.

\paragraph{TPS Throughput and NFE.}

While aggressive gating (high $\beta$) yields marginal accuracy gains, it incurs a catastrophic penalty to efficiency. Thresholds above $1.0$ force the gate to fire almost continuously which inflates the number of forward passes (NFE) and decreasing throughput (TPS), as depicted in Figures~\ref{fig:cp_beta_nfe} and~\ref{fig:cp_beta_tps} respectively, but with a direct trade-off with accuracy.

\section{Generalization to Additional Datasets and Backbones}
\label{app:extra_datasets_backbones}

To test whether \method{}'s gains extend beyond code generation and math reasoning, we evaluate on TruthfulQA~\cite{lin2021truthfulqa}, an open-ended, non-coding, non-reasoning benchmark spanning 38 real-world categories that targets factual accuracy and hallucination resistance. Table~\ref{tab:truthfulqa} reports results on Dream-7B and LLaDA-1.5 across all three base decoders. \method{} improves accuracy in every configuration, with the largest speedups on top of Confidence decoding (5.13$\times$ on Dream-7B, 6.81$\times$ on LLaDA-1.5), since Confidence's largely sequential commit pattern on this dataset leaves the most headroom for \method{}'s anchors.

We additionally evaluate a third dLLM backbone, SDAR-8B~\cite{cheng2026sdar}, on HumanEval and TruthfulQA (MBPP and GSM8K runs were still in progress at submission time). Table~\ref{tab:sdar8b} shows the same pattern observed on Dream and LLaDA: \method{} improves accuracy, throughput, or both across every decoder, indicating the anchor-selection mechanism generalizes across backbone families rather than exploiting properties specific to LLaDA or Dream.